\documentclass[10pt,journal,compsoc]{IEEEtran}

\usepackage{color,xcolor}
\usepackage{epsfig}
\usepackage{graphicx}
\usepackage{subfiles}

\usepackage{adjustbox}
\usepackage{array}
\usepackage{booktabs}
\usepackage{colortbl}
\usepackage{float,wrapfig}
\usepackage{stfloats}
\usepackage{hhline}
\usepackage{multirow}
\usepackage{subcaption} 
\usepackage[labelfont={bf},labelsep={period},font={small}]{caption}

\let\llncssubparagraph\subparagraph
\let\subparagraph\paragraph
\usepackage[compact]{titlesec}
\let\subparagraph\llncssubparagraph

\usepackage{amsmath,amsfonts,amssymb}
\usepackage{bm}
\usepackage{nicefrac}
\usepackage{microtype}

\usepackage{changepage}
\usepackage{extramarks}
\usepackage{fancyhdr}
\usepackage{lastpage}
\usepackage{setspace}
\usepackage{soul}
\usepackage{xspace}

\usepackage{url}

\usepackage{algorithm}
\usepackage{algpseudocode}
\usepackage{enumerate}
\usepackage{times}

\usepackage{threeparttable}
\usepackage{bbding}
\usepackage{makecell}

\usepackage[pagebackref=true,breaklinks=true,letterpaper=true,bookmarks=false]{hyperref}
\usepackage{pbox}
\usepackage{footnote}
\usepackage{tablefootnote}

\usepackage{paralist}

\usepackage{enumitem}
\setitemize{noitemsep,topsep=0pt,parsep=0pt,partopsep=0pt}
\usepackage{paralist}

\makeatletter
\newcommand\footnoteref[1]{\protected@xdef\@thefnmark{\ref{#1}}\@footnotemark}
\makeatother

\makeatletter
\DeclareRobustCommand\onedot{\futurelet\@let@token\@onedot}
\def\@onedot{\ifx\@let@token.\else.\null\fi\xspace}

 \def\vs{\emph{vs}\onedot}

\makeatother

\ifCLASSOPTIONcompsoc
  \usepackage[nocompress]{cite}
\else
  \usepackage{cite}
\fi

\ifCLASSINFOpdf
\else
\fi

\begin{document}

\title{UniFormer: Unifying Convolution and Self-attention for Visual Recognition}

\author{
Kunchang Li, Yali Wang, Junhao Zhang,
Peng Gao, Guanglu Song, \\
Yu Liu,
Hongsheng Li and Yu Qiao
\IEEEcompsocitemizethanks{
\IEEEcompsocthanksitem This work was supported in part by the National Key R\&D Program of China 2022ZD0160505, the Joint Lab of CAS-HK, the National Natural Science Foundation of China under Grant 62272450, the Shenzhen Research Program RCJC20200714114557087, and in part by the Youth Innovation Promotion Association of Chinese Academy of Sciences 2020355. (Kunchang Li and Yali Wang are equally-contributed authors. Yu Qiao is the corresponding author.)
\IEEEcompsocthanksitem Kunchang Li is with ShenZhen Key Lab of Computer Vision and Pattern Recognition, Shenzhen Institute of Advanced Technology, Chinese Academy of Sciences, Shenzhen 518055, China. University of Chinese Academy of Sciences, Beijing 100049, China. (e-mail: kc.li@siat.ac.cn)
\IEEEcompsocthanksitem Yali Wang and Yu Qiao are with ShenZhen Key Lab of Computer Vision and Pattern Recognition, Shenzhen Institute of Advanced Technology, Chinese Academy of Sciences, Shenzhen 518055, China. Shanghai Artificial Intelligence Laboratory, Shanghai 200232, China. (e-mail: yl.wang@siat.ac.cn, qiaoyu@pjlab.org.cn)
\IEEEcompsocthanksitem Junhao Zhang is with National University of Singapore, Peng Gao is with Shanghai Artificial Intelligence Laboratory, Guanglu Song and Yu Liu are with SenseTime Research and Hongsheng Li is with the Chinese University of Hong Kong.
}
}

\markboth{}
{Li \MakeLowercase{\textit{et al.}}: UniFormer: Unifying Convolution and Self-attention for Visual Recognition}

\IEEEtitleabstractindextext{%
\begin{abstract}

It is a challenging task to learn discriminative representation from images and videos,
due to large local redundancy and complex global dependency in these visual data.
Convolution neural networks (CNNs) and vision transformers (ViTs) have been two dominant frameworks in the past few years.
Though CNNs can efficiently decrease local redundancy by convolution within a small neighborhood,
the limited receptive field makes it hard to capture global dependency.
Alternatively,
ViTs can effectively capture long-range dependency via self-attention,
while blind similarity comparisons among all the tokens lead to high redundancy.
To resolve these problems, 
we propose a novel Unified transFormer (UniFormer),
which can seamlessly integrate the merits of convolution and self-attention in a concise transformer format.
Different from the typical transformer blocks,
the relation aggregators in our UniFormer block are equipped with local and global token affinity respectively in shallow and deep layers,
allowing tackling both redundancy and dependency for efficient and effective representation learning.
Finally,
we flexibly stack our blocks into a new powerful backbone,
and adopt it for various vision tasks from image to video domain, from classification to dense prediction.
Without any extra training data,
our UniFormer achieves \textbf{86.3} top-1 accuracy on ImageNet-1K classification task.
With only ImageNet-1K pre-training,
it can simply achieve state-of-the-art performance in a broad range of downstream tasks.
It obtains \textbf{82.9/84.8} top-1 accuracy on Kinetics-400/600, 
\textbf{60.9/71.2} top-1 accuracy on Something-Something V1/V2 video classification tasks,
\textbf{53.8} box AP and \textbf{46.4} mask AP on COCO object detection task,
\textbf{50.8} mIoU on ADE20K semantic segmentation task, 
and \textbf{77.4} AP on COCO pose estimation task. 
Moreover,
we build an efficient UniFormer with a concise hourglass design of token shrinking and recovering,
which achieves \textbf{2-4}$\bm{\times}$ higher throughput than the recent lightweight models.
Code is available at \href{https://github.com/Sense-X/UniFormer}{https://github.com/Sense-X/UniFormer}.

\end{abstract}

}

\maketitle

\IEEEdisplaynontitleabstractindextext

%
\IEEEpeerreviewmaketitle

\IEEEraisesectionheading{\section{Introduction}
\label{sec:introduction}}

\IEEEPARstart{R}{epresentation} learning is a fundamental research topic for visual recognition \cite{swin,vit}.
Basically,
we confront two distinct challenges that exist in visual data such as images and videos. 
On one hand,
the local redundancy is large,
e.g.,
visual content in a local region (space, time or space-time) tends to be similar.
Such locality often introduces inefficient computation.
On the other hand,
the global dependency is complex,
e.g.,
targets in different regions have dynamic relations.
Such long-range interaction often causes ineffective learning.

\begin{figure}[htb]
    \centering
    \includegraphics[width=0.99\linewidth]{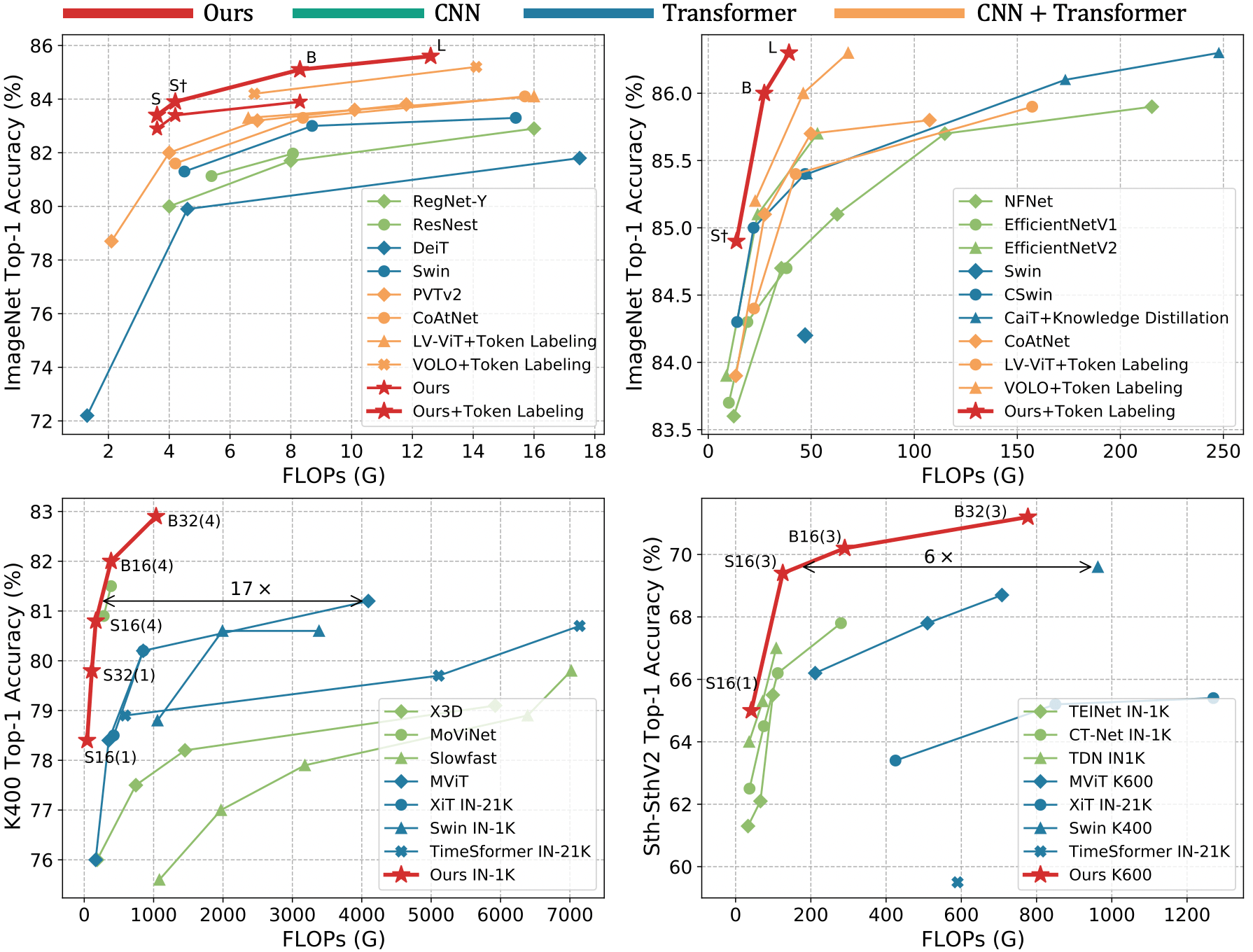}
    \vspace{-0.2cm}
    \caption{\textbf{Accuracy vs. GFLOPs per input}.
    Top: Image Classification (Left/Right: ImageNet with resolution of 224$\times$224/384$\times$384).
    Bottom: Video Classification (Left/Right: K400/Sth-SthV2). 
    `(4)' and `(3)' mean we test UniFormer with 4 clips and 3 crops respectively (more testing details can be found in Section \ref{ablation_studies}).
    Our UniFormer achieves the best balance between accuracy and computation on all the datasets.}
    \label{fig:total_res}
    \vspace{-0.5cm}
\end{figure}

To tackle such difficulties,
researchers have proposed a number of powerful models in visual recognition \cite{resnext,resnest,mvit,morphmlp}.
In particular,
the mainstream backbones are Convolution Neural Networks (CNNs) \cite{resnet,mobilenetv1,slowfast} and Vision Transformers (ViTs) \cite{vit,cait},
where convolution and self-attention are the key operations in these two structures.
Unfortunately, 
each of these operations mainly addresses one aforementioned challenge while ignoring the other.
For example,
the convolution operation is good at reducing local redundancy and avoiding unnecessary computation,
by aggregating each pixel with context from a small neighborhood (e.g, 3$\times$3 or 3$\times$3$\times$3).
However,
the limited receptive field makes convolution suffer from difficulty in learning global dependency \cite{non_local,smallbig}.
Alternatively,
self-attention has been recently highlighted in the ViTs.
By similarity comparison among visual tokens,
it exhibits the strong capacity of learning global dependency in both images \cite{vit,swin} and videos \cite{timesformer,vivit,video_swin}.
Nevertheless,
we observe that ViTs are often inefficient to encode local features in the shallow layers.

\begin{figure*}[tp]
    \vspace{-0.2cm}
    \begin{minipage}[t]{1\textwidth}
        \centering
        \resizebox{0.85\textwidth}{!}{
            \includegraphics[width=0.99\textwidth]{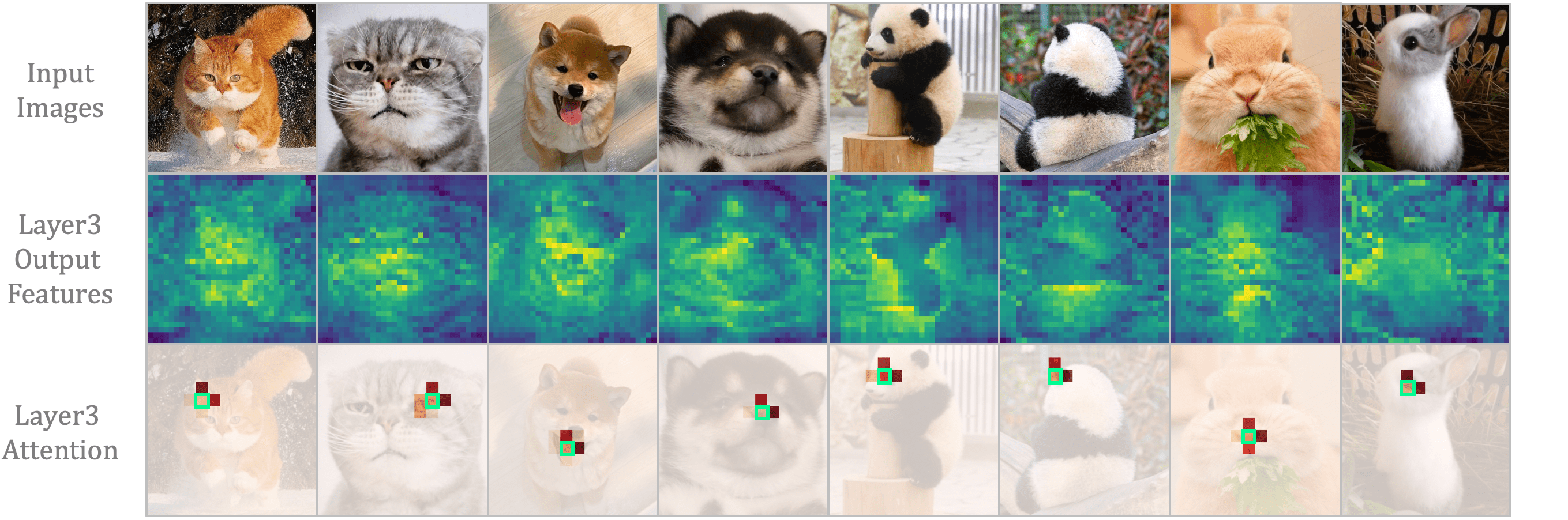}
        }
        \vspace{-0.18cm}
        \subcaption{DeiT.}
        \label{fig:motivation_image}
    \end{minipage}
    \begin{minipage}[t]{1\textwidth}
        \centering
        \resizebox{0.85\textwidth}{!}{
            \includegraphics[width=0.99\textwidth]{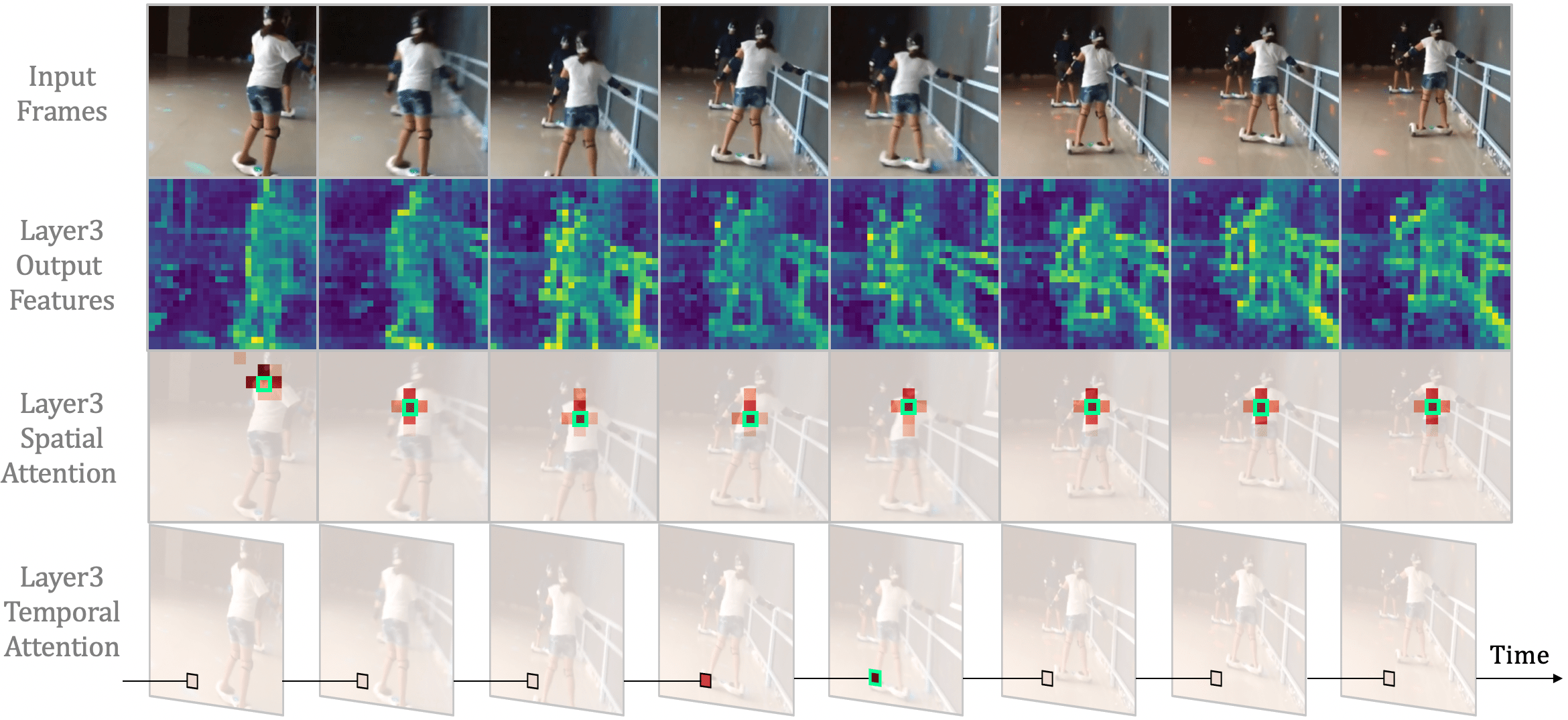}
        }
        \vspace{-0.18cm}
        \subcaption{TimeSformer.}
        \label{fig:motivation_video}
    \end{minipage}
    \vspace{-0.2cm}
    \caption{\textbf{Visualization of vision transformers.}
    We take the well-known Vision Transformers (ViTs) in both image and video domains (i.e., DeiT \cite{deit} and TimeSformer \cite{timesformer}) for illustration,
    where we respectively show the feature maps, spatial and temporal attention maps from the 3rd layer of these ViTs.
    We find that, such ViTs learns local representations with redundant global attention. For an anchor token (green box), spatial/temporal attention compares it with all the contextual tokens for aggregation, while only its neighboring tokens (boxes filled with red color) actually work. Hence, ViTs spend large computation on encoding very local visual representations with global self-attention.}
    \label{fig:motivation}
    \vspace{-0.3cm}
\end{figure*}

We take the well-known ViTs in the image and video domains (i.e.,
DeiT \cite{deit} and TimeSformer \cite{timesformer}) as examples,
and visualize their attention maps in the shallow layer.
As shown in Figure \ref{fig:motivation},
both ViTs indeed capture detailed visual features in the shallow layer,
while spatial and temporal attention are redundant.
One can easily see that,
given an anchor token,
spatial attention largely concentrates on the tokens in the local region (mostly 3$\times$3),
and learns little from the rest tokens in this image.
Similarly, 
temporal attention mainly aggregates the tokens in the adjacent frames, 
while losing sight of the rest tokens in the distant frames. 
However,
such local focus is obtained by global comparison among all the tokens in space and time. 
Clearly,
this redundant attention manner brings large and unnecessary computation burden, 
thus deteriorating the computation-accuracy balance in ViTs (Figure \ref{fig:total_res}).

Based on these discussions,
we propose a novel Unified transFormer (UniFormer) in this work.
It flexibly unifies convolution and self-attention in a concise transformer format,
which can tackle both local redundancy and global dependency for effective and efficient visual recognition.
Specifically,
our UniFormer block consists of three key modules,
i.e.,
Dynamic Position Embedding (DPE), 
Multi-Head Relation Aggregator (MHRA), 
and 
Feed-Forward Network (FFN).
The distinct design of the relation aggregator is the key difference between our UniFormer and the previous CNNs and ViTs.
In the shallow layers, 
our relation aggregator captures local token affinity with a small learnable parameter matrix,
which inherits the convolution style that can largely reduce computation redundancy by context aggregation in the local region.
In the deep layers,
our relation aggregator learns global token affinity with token similarity comparison, 
which inherits the self-attention style that can adaptively build long-range dependency from distant regions or frames. 
Via progressively stacking local and global UniFormer blocks in a hierarchical manner,
we can flexibly integrate their cooperative power to promote representation learning.
Finally,
we provide a generic and powerful backbone for visual recognition
and successfully address various downstream vision tasks with simple and elaborate adaptations.
Additionally,
we further introduce the lightweight design for UniFormer,
which can achieve a preferable accuracy-throughout balance,
by a concise hourglass style of token shrinking and recovering.

Extensive experiments demonstrate the strong performance of our UniFormer on a broad range of vision tasks,
including image classification,
video classification,
object detection,
instance segmentation,
semantic segmentation and pose estimation.
Without any extra training data,
UniFomrer-L achieves \textbf{86.3} top-1 accuracy on ImageNet-1K.
Moreover,
with only ImageNet-1K pre-training,
UniFormer-B achieves \textbf{82.9/84.8} top-1 accuracy on Kinetics-400/Kinetics-600,
\textbf{60.9} and \textbf{71.2} top-1 accuracy on Something-Something V1\&V2,
\textbf{53.8} box AP and \textbf{46.4} mask AP on the COCO detection task,
\textbf{50.8} mIoU on the ADE20K semantic segmentation task,
and \textbf{77.4} AP on the COCO pose estimation task.
Finally,
our efficient UniFormer with a concise hourglass design can achieve \textbf{2-4}$\bm{\times}$ higher throughput than the recent lightweight models.

\begin{figure*}[tp]
    \centering
    \vspace{-0.3cm}
    \includegraphics[width=0.9\textwidth]{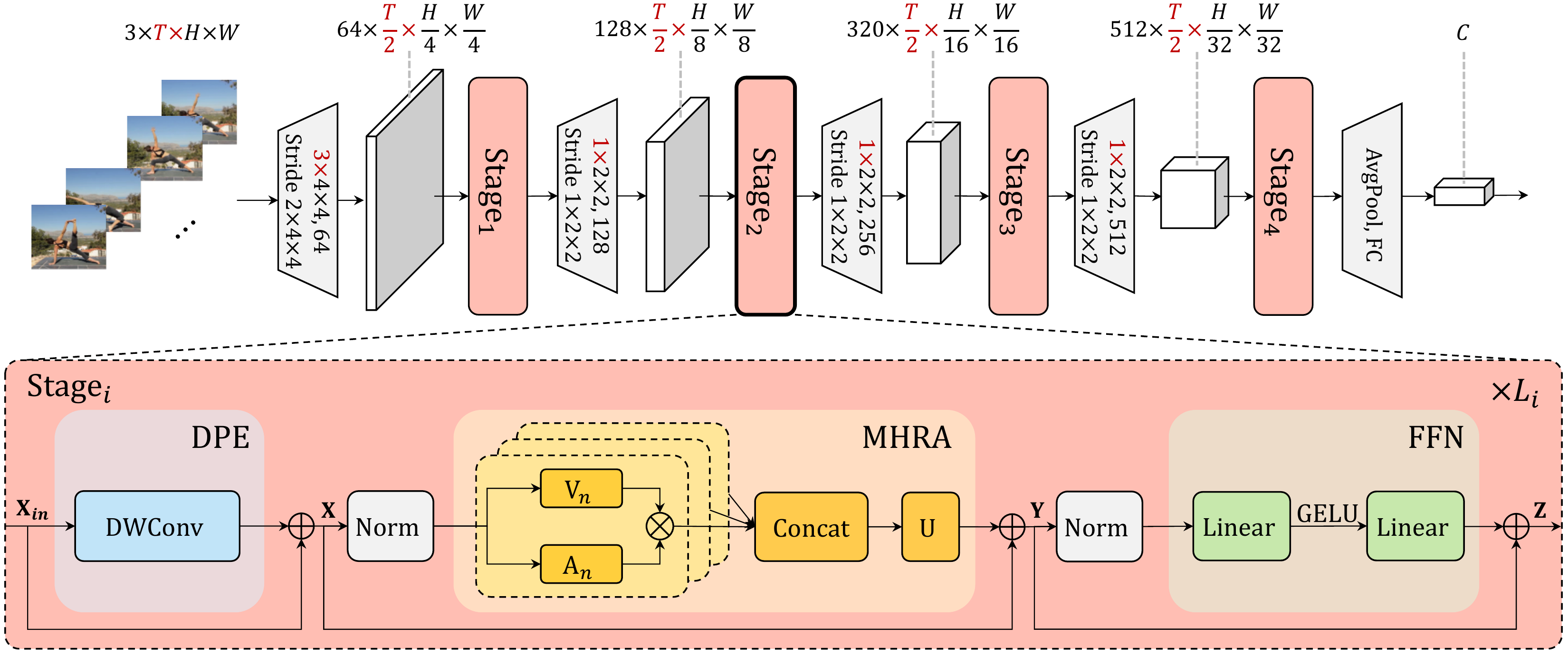}
    \vspace{-0.2cm}
    \caption{\textbf{Unified transFormer (UniFormer).} 
    A UniFormer block consists of three key modules, i.e., Dynamic Position Embedding (DPE), Multi-Head Relation Aggregrator (MHRA), and Feed Forward Network (FFN). 
    The dimensions highlighted in red only exist for the video input,
    while all of them are equal to one for image input.
    More detailed explanations can be found in Section \ref{method}.}
    \label{fig:framework}
    \vspace{-0.3cm}
\end{figure*}

\section{Related Work}

\subsection{Convolution Neural Networks (CNNs)}
In the past few years,
the development of computer vision has been mainly driven by convolutional neural networks (CNNs).
Beginning with the classical AlexNet \cite{alexnet},
many powerful CNN networks have been proposed \cite{vgg,inceptionv1,resnet,resnext,densenet,mobilenetv1,shufflenet,efficientnet} 
and achieved remarkable performance in various tasks of image understanding \cite{imagenet,coco,ade20k,maskrcnn,c-maskrcnn,fpn,upernet,hrnet}.
Recently,
due to the fact that video has gradually become one main data resource in many realistic applications,
researchers have attempted to apply CNNs in the video domain.
Naturally,
one can adapt 2D convolution as 3D one,
by temporal dimension extension \cite{c3d}.
However,
3D CNNs often suffer from difficult optimization problem and large computation cost.
To resolve these issues,
the prior works try to inflate the pre-trained 2D convolution kernels for better optimization \cite{i3d} and factorize 3D convolution kernels in different dimensions to reduce complexity \cite{r(2+1)d,p3d,csn,x3d,slowfast,ct_net}.
Besides,
many recent studies of video understanding \cite{stm,tsm,smallbig,tea} focus on adapting vanilla 2D CNNs with elaborated temporal modeling modules,
such as 
temporal shift \cite{tsm,gst},
motion enhancement \cite{stm,tea,tei},
and spatiotemporal excitation \cite{smallbig,ct_net},
etc.
Unfortunately,
due to the limited reception field,
the traditional convolution struggles to capture long-range dependency even if they are stacked deeper.

\subsection{Vision Transformers (ViTs)}
To capture long-term dependencies,
Vision Transformer (ViT) has been proposed \cite{vit}.
With the inspiration of Transformer architectures in NLP \cite{transformer},
ViT treats image as a number of visual tokens and leverages attention to encode token relations for representation learning.
However, 
vanilla ViT depends on sufficient training data and careful data augmentation.
To tackle these problems, 
several approaches have been developed by 
improved patch embedding \cite{tokenpose},
data-efficient training \cite{deit},
efficient self-attention \cite{swin,cswin,focal},
and multi-scale architectures \cite{pvt,mvit,pvtv2}.
These works successfully boost the performance of ViT on various image tasks \cite{detr,d_detr,segformer,maskformer,TrSeg,tokenpose,transpose,hrformer,TransReID,PreTrainedIP,swinir}.
Recently,
researchers have attempted to extend image ViTs for video modeling.
The classical work is TimeSformer \cite{timesformer} by spatial-temporal attention.
Starting from this,
many works propose different variants for spatiotemporal representation learning \cite{timesformer,vivit,mvit,motionformer,x_vit,video_swin},
and subsequently they are adapted to various video understanding tasks 
\cite{TransformerMT,SSTVOS,MDMMT,MultimodalTF,VideoST}.
Although these works demonstrate the outstanding ability of ViTs to learn long-term token relations,
the self-attention mechanism requires costly token-to-token comparisons \cite{vit}.
Hence, 
it is often inefficient to encode low-level features, as shown in Figure \ref{fig:motivation}.
Though Video Swin \cite{video_swin} advocates an inductive bias of locality with shift window,
window-based self-attention is still less efficient than local convolution \cite{resnet}  when encoding low-level features.
Moreover, 
the shifted window should be carefully configured.

\subsection{Combination of CNNs and ViTs}
To bridge the gap between CNNs and ViTs,
researchers have tried to take advantage of them to build stronger vision backbones for image understanding,
by
adding convolutional patch stem for fast convergence \cite{ceit,early_conv},
introducing convolutional position embedding \cite{cpe,cswin},
inserting depthwise convolution into feed-forward network \cite{ceit,hrformer},
utilizing convolutional projection in self-attention \cite{cvt},
and combining MBConv \cite{mobilenetv2} with Transformer \cite{coatnet}.
As for video understanding,
the combination is also straightforward,
i.e.,
one can insert self-attention as global attention \cite{non_local},
and/or use convolution as patch stem \cite{convtransformer}.
However,
all these approaches ignore inherent relations between convolution and self-attention,
leading to inferior local and/or global token relation learning.
Several recent works have demonstrated that self-attention operates similarly to convolution  \cite{stand_alone,relation_sa}.
But they suggest replacing convolution instead of combining them together.
Differently,
our UniFormer unifies convolution and self-attention in the transformer style,
which can effectively learn local and global token relation, 
and achieve better accuracy-computation trade-offs on all the vision tasks from image to the video domain.

\subsection{Lightweight CNNs and ViTs}
In many practical applications,
the running platforms usually lack enough computility.
Hence,
a series of lightweight CNNs are proposed to satisfy such on-device requirements.
For example,
the classical MobileNets \cite{mobilenetv1,mobilenetv2,mobilenetv3} adopt depthwise separable convolution in well-organized efficient ResNet.
ShuffleNets \cite{shufflenet,shufflenetv2} leverage channel shuffle for computation reduction.
EfficientNets \cite{efficientnet,efficientv2}  further scale model with neural architecture search at depth, width, and resolution.
However,
such lightweight design has not been fully investigated in ViTs.
Two recent works are proposed by
designing transformers as convolutions (i.e., MobielViT\cite{mobilevit}),
and
introducing a parallel architecture of MobileNet and ViT (i.e., MobileFormer\cite{mobileformer}).
But both works ignore the inference speed (e.g., throughout).
Hence,
the prior efficient CNNs are still the better choice.
To bridge this gap,
we build a lightweight UniFormer by token shrinking and recovering in Section \ref{towards_efficient_uniformer}.
\section{Method}
\label{method}
In this section, 
we introduce the proposed UniFormer in detail.
First,
we describe the overview of our UniFormer block.
Then,
we explain its key modules such as multi-head relation aggregator and dynamic position embedding. 
Moreover,
we discuss the distinct relations between our UniFormer and existing CNNs/ViTs, 
showing its preferable design for accuracy-computation balance.

\subsection{Overview}
Figure \ref{fig:framework} shows our Unified transFormer (UniFormer).
For simple description,
we take a video with $T$ frames as an example and an image input can be seen as a video with a single frame.
Hence,
the dimensions highlighted in red only exit for the video input,
while all of them are equal to one for image input.
Our UniFormer is a basic transformer format,
while we elaborately design it to tackle computational redundancy and capture complex dependency. 

Specifically,
our UniFormer block consists of three key modules:
Dynamic Position Embedding (DPE),
Multi-Head Relation Aggregator (MHRA) and Feed-Forward Network (FFN):
\begin{align}
\begin{split}
    \mathbf{X} ={}&
        {\rm DPE}
        \left(
            \mathbf{X}_{in}
        \right) 
        + \mathbf{X}_{in}, \label{dpe}
\end{split}\\
\begin{split}
    \mathbf{Y} ={}&
        {\rm MHRA}
        \left(
            {\rm Norm}
            \left(
                \mathbf{X}
            \right)
        \right) 
        + \mathbf{X}, \label{mhr_r}
\end{split}\\ 
    \mathbf{Z} ={}&
        {\rm FFN}
        \left(
            {\rm Norm}
            \left(
                \mathbf{Y}
            \right)
        \right) 
        + \mathbf{Y}.  \label{ffn}
\end{align}
Considering the input token tensor $\mathbf{X}_{in}\in \mathbb{R}^{C\times T\times H\times W}$ ($T$$=$$1$ for an image input),
we first introduce DPE to dynamically integrate position information into all the tokens (Eq. \ref{dpe}).
It is friendly to arbitrary input resolution and makes good use of token order for better visual recognition.
Then,
we use MHRA to enhance each token by exploiting its contextual tokens with relation learning (Eq. \ref{mhr_r}).
Via flexibly designing the token affinity in the shallow and deep layers,
our MHRA can smartly unify convolution and self-attention to reduce local redundancy and learn global dependency.
Finally,
we add FFN like traditional ViTs \cite{vit}, 
which consists of two linear layers and one non-linear function,
i.e., 
GELU (Eq. \ref{ffn}).
The channel number is first expanded by the ratio of 4 and then recovered,
thus each token will be enhanced individually.

\subsection{Multi-Head Relation Aggregator}
\label{mhra_section}
As analyzed before,
the traditional CNNs and ViTs focus on addressing either local redundancy or global dependency,
leading to unsatisfactory accuracy or/and unnecessary computation.
To overcome these difficulties,
we introduce a generic Relation Aggregator (RA), 
which elegantly unifies convolution and self-attention for token relation learning.
It can achieve efficient and effective representation learning by designing local and global token affinity in the shallow and deep layers respectively.
Specifically,
MHRA exploits token relationships in a multi-head style:
\begin{align}
\begin{split}
     {\rm R}_n(\mathbf{X}) ={}&
        {\rm A}_n{\rm V}_n(\mathbf{X}), \label{a}
\end{split} \\
    {\rm MHRA}(\mathbf{X}) ={}&
        {\rm Concat}(
            {\rm R}_1(\mathbf{X}); 
            {\rm R}_2(\mathbf{X}); 
            \cdots; 
            {\rm R}_N(\mathbf{X})
        )
        \mathbf{U}. \label{mhr}
\end{align}
Given the input tensor $\mathbf{X}\in \mathbb{R}^{C\times T\times H\times W}$,
we first reshape it to a sequence of tokens $\mathbf{X}\in \mathbb{R}^{L\times C}$ with the length of $L$$=$$T$$\times$$H$$\times$$W$.
Moreover,
${\rm R}_n(\cdot)$ refers to RA in the $n$-th head and $\mathbf{U}\in \mathbb{R}^{C\times C}$ is a learnable parameter matrix to integrate $N$ heads.
Each RA consists of token context encoding and token affinity learning.
We apply a linear transformation to encode the original tokens into contextual tokens ${\rm V}_n(\mathbf{X}) \in \mathbb{R}^{L\times \frac{C}{N}}$.
Subsequently,
RA can summarize context with the guidance of token affinity ${\rm A}_n \in \mathbb{R}^{L\times L}$.
We will describe how to learn the specific ${\rm A}_n$ in the following.

\subsubsection{Local MHRA}
As shown in Figure \ref{fig:motivation},
though the previous ViTs compare similarities among all the tokens, 
they finally learn local representations. 
Such redundant self-attention design brings large computation cost in the shallow layers.
Based on it,
we suggest learning token affinity in a small neighborhood,
which coincidentally shares a similar insight with the design of a convolution filter.
Hence,
we propose to represent local affinity as a learnable parameter matrix in the shallow layers.
Concretely,
given an anchor token $\mathbf{X}_{i}$,
our local RA learns the affinity between this token and other tokens in the small neighborhood $\Omega_{i}^{t\times h\times w}$ ($t$$=$$1$ for an image input):
\begin{align}
    {\rm A}_n^{local}(\mathbf{X}_{i}, \mathbf{X}_{j}) ={}&
        a_{n}^{i-j},\ \ where\ \ j\in \Omega_{i}^{t\times h\times w}, \label{local_a}
\end{align}
where $a_n\in \mathbb{R}^{t\times h\times w}$ is the learnable parameter, 
and $\mathbf{X}_j$ refers to any neighbor token in $\Omega_{i}^{t\times h\times w}$. 
$(i-j)$ denotes the relative position between token $i$ and $j$.
Note that,
visual content between adjacent tokens varies subtly in the shallow layers, 
since the receptive field of tokens is small.
In this case, 
it is not necessary to make token affinity dynamic in these layers. 
Hence, 
we use a learnable parameter matrix to describe local token affinity, 
which simply depends on the relative position between tokens.

\textbf{Comparison to Convolution Block.}
Interestingly,
we find that our local MHRA can be interpreted as a generic extension of MobileNet block \cite{mobilenetv2,csn,x3d}.
Firstly,
the linear transformation ${\rm V}(\cdot)$ in Eq. \ref{a} is equivalent to a pointwise convolution (PWConv),
where each head is corresponding to an output feature channel ${\rm V}_n({\rm X})$.
Furthermore,
our local token affinity ${\rm A}_n^{local}$ can be instantiated as the parameter matrix that operated on each output channel (or head) ${\rm V}_n(\mathbf{X})$,
thus the relation aggregator ${\rm R}_n(\mathbf{X}) = {\rm A}_n^{local}{\rm V}_n(\mathbf{X})$ can be explained as a depthwise convolution (DWConv).
Finally,
the linear matrix $\mathbf{U}$, 
which concatenates and fuses all heads,
can also be seen as a pointwise convolution.
As a result,
such local MHRA can be reformulated with a manner of PWConv-DWConv-PWConv in the MobileNet block.
In our experiments,
we instantiate our local MHRA as such channel-separated convolution,
so that our UniFormer can boost computation efficiency for visual recognition.
Moreover,
different from the MobileNet block, 
our local UniFormer block is designed as a generic transformer format,
i.e.,
it also contains dynamical position encoding (DPE) and feed-forward network (FFN),
besides MHRA.
This unique integration can effectively enhance token representation,
which has not been explored in the previous convolution blocks. 

\subsubsection{Global MHRA}
In the deep layers,
it is important to exploit long-range relation in the broader token space,
which naturally shares a similar insight with the design of self-attention.
Therefore,
we design the token affinity via comparing content similarity among all the tokens:
\begin{align}
    {\rm A}_n^{global}(\mathbf{X}_{i}, \mathbf{X}_{j}) ={}&
        \frac{
            e^{Q_n(\mathbf{X}_{i})^{T}K_n(\mathbf{X}_{j})}
        }{
            \sum_{j'\in \Omega_{T\times H\times W}}
            e^{Q_n(\mathbf{X}_{i})^{T}
            K_n(\mathbf{X}_{j'})}
        }, \label{global_a}
\end{align}
where $\mathbf{X}_{j}$ can be any token in the global tube with a size of $T$$\times$$H$$\times$$W$ ($T$$=$$1$ for an image input),
while $Q_n(\cdot)$ and $K_n(\cdot)$ are two different linear transformations.

\textbf{Comparison to Transformer Block}.
Our global MHRA ${\rm A}_n^{global}$ (Eq. \ref{global_a}) can be instantiated as a spatiotemporal self attention,
where
${\rm Q}_n(\cdot)$, ${\rm K}_n(\cdot)$ and ${\rm V}_n(\cdot)$ become Query, Key and Value in ViT \cite{vit}.
Hence,
it can effectively learn long-range dependency.
However,
our global UniFormer block is different from the previous ViT blocks.
First,
most video transformers divide spatial and temporal attention in the video domain \cite{timesformer,vivit,stam},
in order to reduce the dot-product computation in token similarity comparison.
But such an operation inevitably deteriorates the spatiotemporal relation among tokens. 
In contrast,
our global UniFormer block jointly encodes spatiotemporal token relation to generate more discriminative video representation for recognition.
Since our local UniFormer block largely saves computation of token comparison in the shallow layers,
the overall model can achieve a preferable computation-accuracy balance.
Second,
instead of absolute position embedding \cite{vit,transformer},
we adopt dynamic position embedding (DPE) in our UniFormer.
It is in convolution style (see the next section),
which can overcome permutation-invariance and be friendly to different input lengths of visual tokens.

\begin{table}[tp]
	\centering
    \setlength\tabcolsep{2pt}
    \resizebox{\linewidth}{!}{
        \begin{tabular}[t]{c|c|c|c|c|c}
            \Xhline{1.0pt}
            Model & Type & \#Blocks & \#Channels & \#Param. & FLOPs \\ 
            \Xhline{1.0pt}
            Small & [L, L, G, G] & [3, 4, 8, 3] & [64, 128, 320, 512] & 21.5M & 3.6G \\
            Base & [L, L, G, G] & [5, 8, 20, 7] & [64, 128, 320, 512] & 50.3M & 8.3G \\
            Large & [L, L, G, G] & [5, 10, 24, 7] & [128, 192, 448, 640] & 100M & 12.6G \\
            \Xhline{1.0pt}
        \end{tabular}
    }
    \vspace{-0.2cm} 
    \caption{\textbf{Backbones for image classification.} `L' and `G' refer to our local and global UniFormer blocks respectively.
    The FLOPs are measured at resolution 224$\times$224.
    }
    \label{model_detail}
    \vspace{-0.4cm}
\end{table}  

\subsection{Dynamic Position Embedding}
\label{dpe_section}
The position information is an important clue to describe visual representation.
Previously,
most ViTs encode such information by absolute or relative position embedding \cite{vit,pvt,swin}.
However,
absolute position embedding has to be interpolated for various input sizes with fine-tuning \cite{deit,cait},
while
relative position embedding does not work well due to the modification of self-attention \cite{cpe}.
To improve flexibility,
convolutional position embedding has been recently proposed \cite{twins,cswin}.
In particular,
conditional position encoding (CPE) \cite{cpe} can implicitly encode position information via convolution operators,
which unlocks Transformer to process arbitrary input size and promotes recognition performance.
Due to its plug-and-play property,
we flexibly adopt it as our Dynamical Position Embedding (DPE) in the UniFormer:
\begin{align}
    {\rm DPE}(\mathbf{X}_{in}) = {\rm DWConv}(\mathbf{X}_{in}),
\end{align}
where ${\rm DWConv}$ refers to depthwise convolution with zero paddings. 
We choose such a design as our DPE based on the following reasons.
First,
depthwise convolution is friendly to arbitrary input shapes,
e.g.,
it is straightforward to use its spatiotemporal version to encode 3D position information in videos.
Second,
depthwise convolution is light-weight,
which is an important factor for computation-accuracy balance.
Finally,
we add extra zero paddings,
since it can help tokens be aware of their absolute positions by querying their neighbors progressively \cite{cpe}.

\section{Framework}
In the section,
we mainly develop visual frameworks for various downstream tasks.
Specifically,
we first develop a number of visual backbones for image classification,
by hierarchically stacking our local and global UniFormer blocks with consideration of computation-accuracy balance.
Then,
we extend the above backbones to tackle other representative vision tasks,
including video classification and dense prediction (i.e.,
object detection, 
semantic segmentation and human pose estimation). 
Such generality and flexibility of our UniFormer demonstrate its valuable potential for computer vision research and beyond.

\begin{figure}[tp]
    \centering
    \vspace{-0.2cm}
    \includegraphics[width=0.95\linewidth]{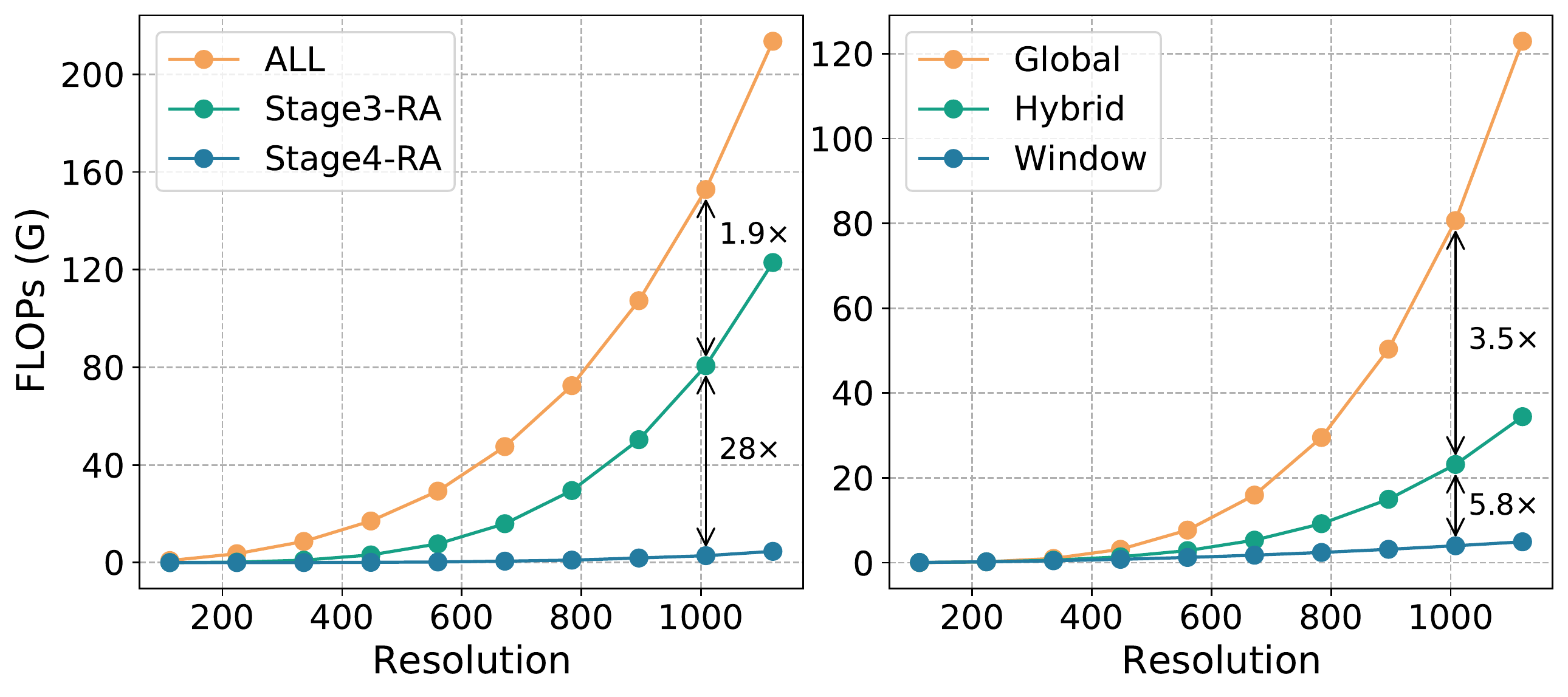}
    \vspace{-0.2cm}
    \caption{\textbf{FLOPs vs. Resolution.} \textbf{Left:} Total FLOPs and the FLOPs of MatMul in RA in Stage3 and Stage4. RA in Stage3 requires more computation for lage resolution. \textbf{Right:} The FLOPs of MatMul in different RA in Stage3. Window and hybrid blocks saves computation.}
    \label{fig:flops}
    \vspace{-0.3cm}
\end{figure}

\subsection{Image Classification}
\label{section_framework}

It is important to progressively learn visual representation for capturing semantics in the image.
Hence,
we build up our backbone with four stages,
as illustrated in Figure \ref{fig:framework}.

More specifically,
we use the local UniFormer blocks in the first two stages to reduce computation redundancy,
while the global UniFormer blocks are utilized in the last two stages to learn long-range token dependency.
For the local UniFormer block,
MHRA is instantiated as PWConv-DWConv-PWConv with local token affinity (Eq. \ref{local_a}),
where the spatial size of DWConv is set to 5$\times$5 for image classification.
For the global UniFormer block,
MHRA is instantiated as multi-head self-attention with global token affinity (Eq. \ref{global_a}),
where the number of attention heads is set to 64.
For both local and global UniFormer blocks,
DPE is instantiated as DWConv with a spatial size of 3$\times$3,
and the expand ratio of FFN is 4.

Additionally,
as suggested in the CNN and ViT literatures \cite{resnet,vit},
we utilize BN \cite{bn} for convolution and LN \cite{ln} for self-attention.
For feature downsampling,
we use the 4$\times$4 convolution with stride 4$\times$4 before the first stage
and the 2$\times$2 convolution with stride 2$\times$2 before other stages.
Besides,
an extra LN is added after each downsampling convolution.
Finally, 
the global average pooling and fully connected layer are applied to output the predictions.
When training models with Token Labeling \cite{lvvit},
we add another fully connected layer for auxiliary loss. 
For various computation requirements,
we design three model variants as shown in Table \ref{model_detail}.

\subsection{Video Classification}
Given our image-based 2D backbones,
one can easily adapt them as 3D backbones for video classification.
Without loss of generality,
we adjust Small and Base models for spatiotemporal modeling.
Specifically,
the model architectures keep the same with four stages,
where
we use the local UniFormer blocks in the first two stages and the global UniFormer blocks in the last two stages.
But differently,
all the 2D convolution filters are changed as 3D ones via filter inflation \cite{i3d}.
Concretely, 
the kernel size of DWConv in DPE and local MHRA are 3$\times$3$\times$3 and 5$\times$5$\times$5 respectively.
Moreover,
we downsample both spatial and temporal dimensions before the first stage.
Hence,
the convolution filter before this stage becomes 3$\times$4$\times$4 with the stride of 2$\times$4$\times$4.
For the other stages,
we just downsample the spatial dimension to decrease the computation cost and maintain high performance.
Hence,
the convolution filters before these stages are 1$\times$2$\times$2 with stride of 1$\times$2$\times$2 .

Note that we use spatiotemporal attention in the global UniFormer blocks for learning token relation jointly in the 3D view.
It is worth mentioning that,
due to the large model sizes,
the previous video transformers \cite{timesformer,vivit} divide spatial and temporal attention to reduce computation and alleviate overfitting,
but such factorization operation inevitably tears spatiotemporal token relations.
In contrast,
our joint spatiotemporal attention can avoid the issue.
Besides,
our local UniFormer blocks largely save computation via 3D DWconv.
Hence, 
our model can achieve effective and efficient video representation learning.

\begin{figure*}[tp]
    \centering
    \vspace{-0.2cm}
    \includegraphics[width=0.99\textwidth]{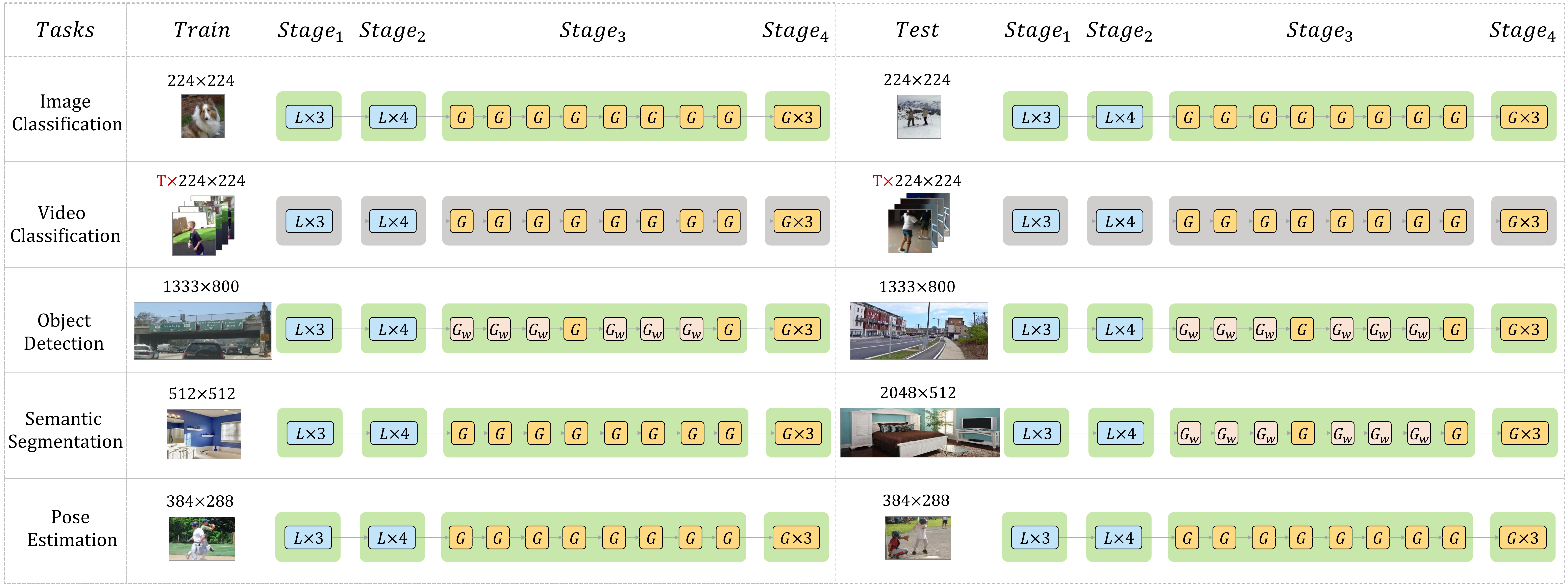}
    \vspace{-0.2cm}
    \caption{\textbf{Training and testing adaption for downstream tasks.} For video classification, we inflate all the 2D convolution filters to 3D ones. For dense prediction, we modify RA in Stage3 for different downstream tasks. `$G_w$' means we apply global MHRA in a predefined window.}
    \label{fig:adaption}
    \vspace{-0.3cm}
\end{figure*}

\subsection{Dense Prediction}
\label{adaption_section}
Dense prediction tasks are necessary to verify the generality of our recognition backbones.
Hence,
we adopt our UniFormer backbones for a number of popular dense tasks such as object detection,
instance segmentation,
semantic segmentation,
and
human pose estimation.
However,
direct usage of our backbone is not suitable because of the high input resolution of most dense prediction tasks,
e.g.,
the size of the image is 1333$\times$800 in the COCO object detection dataset.
Naturally,
feeding such images into our classification backbones would inevitably lead to large computation,
especially when operating self-attention of global UniFormer block in the last two stages.
Taking $h$$\times$$w$ visual tokens as an example,
the MatMul operation in token similarity comparison (Eq. \ref{global_a}) causes ${\mathcal O}(w^2h^2)$ complexity,
which is prohibitive for most dense tasks.

We propose to adjust the global UniFormer block for different downstream tasks.
First,
we analyze the FLOPs of our UniFormer-S under different input resolutions. 
Figure \ref{fig:flops} clearly indicates that Relation Aggregator (RA) in Stage3 occupies large computation.
For example,
for a 1008$\times$1008 image,
the MatMul operation of RA in Stage3 even occupies over 50\% of the total FLOPs,
while the FLOPs in Stage4 is only 1$/$28 of that in Stage3.
Thus we focus on modifying RA in Stage3 for computation reduction.

Inspired by \cite{stand_alone,swin},
we propose to apply our global MHRA in a predefined window (e.g., 14$\times$14),
instead of using it in the entire image with high resolution.
Such operation can effectively cut down computation with the complexity of ${\mathcal O}(whp^2)$,
where
$p$ is the window size. 
However,
it undoubtedly drops model performance,
due to insufficient token interaction. 
To bridge this gap, 
we integrate window and global UniFormer blocks together in Stage3,
where a hybrid group consists of three window blocks and one global block.
In this case,
there are 2/5 hybrid groups in Stage3 of our UniFormer-Small/Base backbones.

Based on this design,
we next introduce the specific backbone settings of various dense tasks,
depending on the input resolution of training and testing images.
For object detection and instance segmentation,
the input images are usually large (e.g., 1333$\times$800),
thus we adopt the hybrid block style in Stage3.
In contrast,
the inputs are relatively small for pose estimation,
such as 384$\times$288,
hence global blocks are still applied in Stage3 for both training and testing.
Specially,
for semantic segmentation,
the testing images are often larger than the training ones.
Therefore,
we utilize the global blocks in Stage3 for training,
while adapting the hybrid blocks in Stage3 for testing.
We use the following simple design.
The window-based block in testing has the same receptive field as the global block in training,
e.g., 32$\times$32.
Such design can maintain training efficiency,
and boost testing performance by keeping consistency with training as much as possible.

\section{Towards Lightweight UniFormer}
\label{towards_efficient_uniformer}

\begin{figure}[tp]
    \centering
    \vspace{-0.2cm}
    \includegraphics[width=0.99\linewidth]{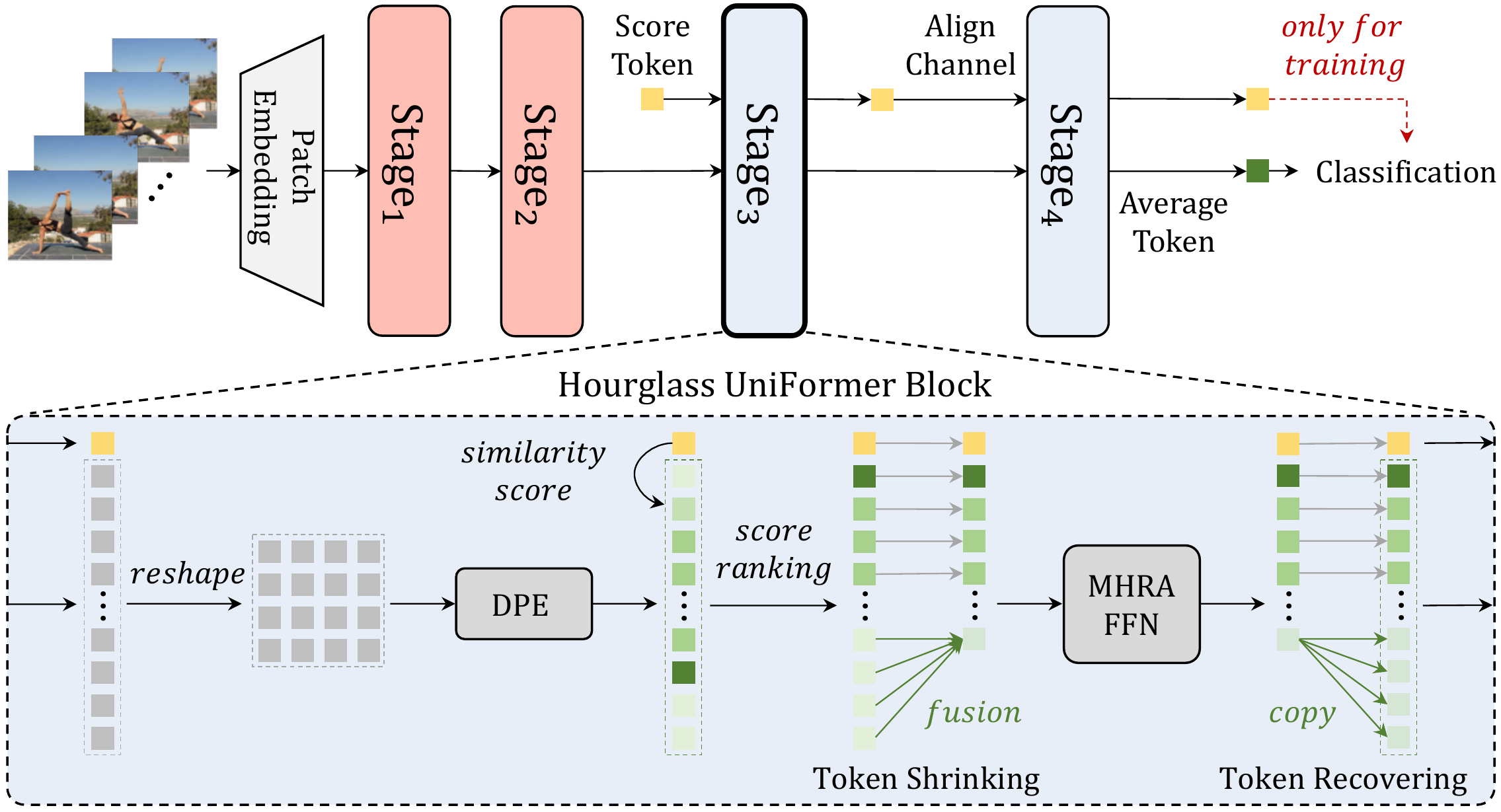}
    \vspace{-0.2cm}
    \caption{\textbf{Lightweight UniFormer.}
    We propose an Hourglass UniFormer block,
    saving computation of MHRA and FFN by adaptive token shrinking and recovering.}
    \label{fig:efficient_uniformer}
    \vspace{-0.4cm}
\end{figure}

Recently,
researchers have tried to combine CNNs with ViTs to design lightweight models.
For example,
MobileFormer\cite{mobileformer} proposes a parallel design of MobileNet\cite{mobilenetv2} and ViT\cite{vit},
and MobileViT\cite{mobilevit} designs transformers as convolutions.
However,
the inference speeds of these works should be further improved.
Hence,
the prior efficient CNNs are still the better choice,
such as EfficientNet \cite{efficientnet} for image tasks and MoViNet\cite{movienet} for video tasks.
To bridge the gap,
we propose a lightweight UniFormer architecture in Figure \ref{fig:efficient_uniformer},
by designing a distinct hourglass UniFormer block.
In this block,
we adaptively leverage token shrinking and recovering,
to achieve a preferable accuracy-throughput balance.


\subsection{Hourglass UniFormer Block}

Since a large computation load lies in token similarity comparison in the global UniFormer block,
we propose an Hourglass UniFormer (H-UniFormer) block to reduce the number of visual tokens involved in global MHRA.
Note that,
the existing token pruning methods \cite{dynamicvit,evit,sit} are infeasible for our UniFormer and those ViTs with convolution \cite{pvtv2,ceit,coatnet}.
The main reason is that,
after pruning,
the rest tokens often maintain a broken spatiotemporal structure,
which makes convolution inapplicable.
To overcome such difficulty,
we propose a concise integration of token shrinking and recovering in our H-UniFormer block.

\textbf{Token Shrinking}.
We introduce a score token $\mathbf{s}\in\mathbb{R}^{C}$ to measure the importance of the visual tokens $\mathbf{X}\in \mathbb{R}^{C\times THW}=\{\mathbf{X}_{1},...,\mathbf{X}_{THW}\}$ after DPE.
Specifically,
we compare similarity (using Eq. \ref{global_a}) between the score token $\mathbf{s}$ and a visual token $\mathbf{X}_{j}$,
and average the similarity values over $N$ attention heads,
\begin{equation}
\mathbf{A}_j= \sum_{n=1}^{N} {\rm A}_n^{global}(\mathbf{s}, \mathbf{X}_j) / N ,
\end{equation}
where $\mathbf{A}_j$ represents the importance of visual token $\mathbf{X}_{j}$,
and 
we use $\mathbf{A}\in \mathbb{R}^{THW}$ to represent the importance vector of all the visual tokens.
For tokens with high values in $\mathbf{A}$,
we consider them to be crucial tokens and keep them.
For tokens with low values,
we consider them to be unimportant tokens.
Hence,
we fuse them as one representative token,
where 
their score values are used as the fusing weights.
By such unimportant token reduction,
we can shrink visual tokens 
from 
$\mathbf{X}\in \mathbb{R}^{C\times THW}$ 
to 
$\mathbf{X}^{S}\in \mathbb{R}^{C\times M}$,
where 
the number of tokens $M \ll THW$.
Subsequently,
the reduced visual tokens are fed into global MHRA and FFN,
for computation saving.
Additionally,
we pass this score token through all the H-UniFormer blocks in Stage3 and Stage4,
and use it to calculate the classification loss when training.
In this case,
the score token is effectively guided by the ground-truth label,
and it thus becomes discriminative to weight token importance.

\textbf{Token Recovering}.
After learning token interactions in global MHRA and FFN,
we replicate the representative token to recover the unimportant tokens.
Thus,
we can maintain the spatiotemporal structure of all the visual tokens,
for effective dynamic position encoding (i.e., convolution) in the next H-UniFormer block.

\begin{table}[tp]
	\centering
    \setlength\tabcolsep{2pt}
    \resizebox{0.99\linewidth}{!}{
    	\begin{tabular}{c|l|cc|cc|c}
    	    \Xhline{1.0pt}
    		\multirow{2}*{Arch.} & \multirow{2}*{Method} & \#Param & FLOPs & Train & Test & ImageNet \\
    		 ~ & ~ & (M) & (G) & Size  & Size & Top-1  \\
    	    \Xhline{1.0pt}
    		\multirow{3}{*}{\rotatebox{90}{CNN}} & RegNetY-4G \cite{regnet} & 21 & 4.0 & 224 & 224 & 80.0 \\
    		~ & EffcientNet-B5 \cite{efficientnet} & 30 & 9.9 & 456 & 456 & 83.6 \\
    		~ & EfficientNetV2-S \cite{efficientv2} & 22 & 8.5 & 384 & 384 & 83.9 \\
    		\hline
    		\multirow{7}{*}{\rotatebox{90}{Trans}} & DeiT-S \cite{deit} & 22 & 4.6 & 224 & 224 & 79.9 \\
    		~ & PVT-S \cite{pvt} & 25 & 3.8 & 224 & 224 & 79.8 \\
    		~ & T2T-14 \cite{t2t} & 22 & 5.2 & 224 & 224 & 80.7 \\
    		~ & Swin-T \cite{swin} & 29 & 4.5 & 224 & 224 & 81.3 \\
    		~ & Focal-T \cite{focal} & 29 & 4.9 & 224 & 224 & 82.2 \\
    		~ & CSwin-T \cite{cswin} & 23 & 4.3 & 224 & 224 & 82.7 \\
    		~ & CSwin-T $\uparrow$384 \cite{cswin} & 23 & 14.0 & 224 & 384 & 84.3 \\
    		\hline
    		\multirow{12}{*}{\rotatebox{90}{CNN+Trans}} & CvT-13 \cite{cvt} &  20 & 4.5 & 224 & 224 & 81.6 \\
    		~ & CvT-13 $\uparrow$384 \cite{cvt} &  20 & 16.3 & 224 & 384 & 83.0 \\
    		~ & CoAtNet-0 \cite{coatnet} &  25 & 4.2 & 224 & 224 & 81.6 \\
    		~ & CoAtNet-0 $\uparrow$384 \cite{coatnet} &  20 & 13.4 & 224 & 384 & 83.9 \\
    		~ & Container \cite{container} &  22 & 8.1 & 224 & 224 & 82.7 \\
    		~ & LV-ViT-S \cite{lvvit} & 26 & 6.6 & 224 & 224 & 83.3 \\
    		~ & LV-ViT-S $\uparrow$384 \cite{lvvit} & 26 & 22.2 & 224 & 384 & 84.4 \\
    		~ & \cellcolor{gray!20}{UniFormer-S} & \cellcolor{gray!20}{22} & \cellcolor{gray!20}{3.6} & \cellcolor{gray!20}{224} & \cellcolor{gray!20}{224} & \cellcolor{gray!20}{82.9} \\
    		~ & \cellcolor{gray!20}{UniFormer-S$^{\star}$} & \cellcolor{gray!20}{22} & \cellcolor{gray!20}{3.6} & \cellcolor{gray!20}{224} & \cellcolor{gray!20}{224} & \cellcolor{gray!20}{83.4} \\
    		~ & \cellcolor{gray!20}{UniFormer-S$^{\star}$ $\uparrow$384} & \cellcolor{gray!20}{22} & \cellcolor{gray!20}{11.9} & \cellcolor{gray!20}{224} & \cellcolor{gray!20}{384} & \cellcolor{gray!20}{84.6} \\
    		~ & \cellcolor{gray!20}{UniFormer-S$\dagger$} & \cellcolor{gray!20}{24} & \cellcolor{gray!20}{4.2} & \cellcolor{gray!20}{224} & \cellcolor{gray!20}{224} & \cellcolor{gray!20}{83.4} \\
    		~ & \cellcolor{gray!20}{UniFormer-S$\dagger^{\star}$} &  \cellcolor{gray!20}{24} & \cellcolor{gray!20}{4.2} & \cellcolor{gray!20}{224} & \cellcolor{gray!20}{224} & \cellcolor{gray!20}{83.9} \\
    		~ & \cellcolor{gray!20}{UniFormer-S$\dagger^{\star}$ $\uparrow$384} &  \cellcolor{gray!20}{24} & \cellcolor{gray!20}{13.7} & \cellcolor{gray!20}{224} & \cellcolor{gray!20}{384} & \cellcolor{gray!20}{\textbf{84.9}} \\
    	    \Xhline{1.0pt}
    		\multirow{3}{*}{\rotatebox{90}{CNN}} & RegNetY-8G \cite{regnet} & 39 & 8.0 & 224 & 224 & 81.7 \\
    		~ & EffcientNet-B7 \cite{efficientnet} & 66 & 39.2 & 600 & 600 & 84.3 \\
    		~ & EfficientNetV2-M \cite{efficientv2} & 54 & 25.0 & 480 & 480 & 85.1 \\
    		\hline
    		\multirow{6}{*}{\rotatebox{90}{Trans}} & PVT-L \cite{pvt} & 61 & 9.8 & 224 & 224 & 81.7 \\
    		~ & T2T-24 \cite{t2t} & 64 & 13.2 & 224 & 224 & 82.2 \\
    		~ & Swin-S \cite{swin} & 50 & 8.7 & 224 & 224 & 83.0 \\
    		~ & Focal-S \cite{focal} & 51 & 9.1 & 224 & 224 & 83.5 \\
    		~ & CSwin-S \cite{cswin} & 35 & 6.9 & 224 & 224 & 83.6 \\
    		~ & CSwin-S $\uparrow$384 \cite{cswin} & 35 & 22.0 & 224 & 384 & 85.0 \\
    		\hline
    		\multirow{8}{*}{\rotatebox{90}{CNN+Trans}} & CvT-21 \cite{cvt} &  32 & 7.1 & 224 & 224 & 82.5 \\
    		~ & CoAtNet-1 \cite{coatnet} &  42 & 8.4 & 224 & 224 & 83.3 \\
    		~ & CoAtNet-1 $\uparrow$384 \cite{coatnet} &  42 & 27.4 & 224 & 384 & 85.1 \\
    		~ & LV-ViT-M \cite{lvvit} & 56 & 16.0 & 224 & 224 & 84.1 \\
    		~ & LV-ViT-M $\uparrow$384 \cite{lvvit} & 56 & 42.2 & 224 & 384 & 85.4 \\
    		~ & \cellcolor{gray!20}{UniFormer-B} &  \cellcolor{gray!20}{50} & \cellcolor{gray!20}{8.3} & \cellcolor{gray!20}{224} & \cellcolor{gray!20}{224} & \cellcolor{gray!20}{83.9} \\
    		~ & \cellcolor{gray!20}{UniFormer-B$^{\star}$} &  \cellcolor{gray!20}{50} & \cellcolor{gray!20}{8.3} & \cellcolor{gray!20}{224} & \cellcolor{gray!20}{224} & \cellcolor{gray!20}{85.1} \\
    		~ & \cellcolor{gray!20}{UniFormer-B$^{\star}$ $\uparrow$384} &  \cellcolor{gray!20}{50} & \cellcolor{gray!20}{27.2} & \cellcolor{gray!20}{224} & \cellcolor{gray!20}{384} & \cellcolor{gray!20}{\textbf{86.0}} \\
    	    \Xhline{1.0pt}
    		\multirow{3}{*}{\rotatebox{90}{CNN}} & RegNetY-16G \cite{regnet} & 84 & 16.0 & 224 & 224 & 82.9 \\
    		~ & EfficientNetV2-L \cite{efficientv2} & 121 & 53 & 480 & 480 & 85.7 \\
    		~ & NFNet-F4 \cite{nfnet} & 316 & 215.3 & 384 & 512 & 85.9 \\
    		\hline
    		\multirow{8}{*}{\rotatebox{90}{Trans}} & DeiT-B \cite{deit} & 86 & 17.5 & 224 & 224 & 81.8 \\
    		~ & Swin-B \cite{swin} & 88 & 15.4 & 224 & 224 & 83.3 \\
    		~ & Swin-B $\uparrow$384 & 88 & 47.0 & 224 & 384 & 84.2 \\
    		~ & Focal-B \cite{swin} & 90 & 16.0 & 224 & 224 & 83.8 \\
    		~ & CSwin-B \cite{cswin} & 78 & 15.0 & 224 & 224 & 84.2 \\
    		~ & CSwin-B $\uparrow$384 \cite{cswin} & 78 & 47.0 & 224 & 384 & 85.4 \\
            ~ & CaiT-S36 $\uparrow$384$\Upsilon$ \cite{cait} & 68 & 48.0 & 224 & 384 & 85.4 \\
    		~ & CaiT-M36 $\uparrow$448$\Upsilon$ \cite{cait} & 271 & 247.8 & 224 & 448 & 86.3 \\
    		\hline
    		\multirow{9}{*}{\rotatebox{90}{CNN+Trans}} & BoTNet-T7 \cite{botnet} &  79 & 19.3 & 256 & 256 & 84.2 \\
    		~ & CoAtNet-3 \cite{coatnet} &  168 & 34.7 & 224 & 224 & 84.5 \\
    		~ & CoAtNet-3 $\uparrow$384 \cite{coatnet} &  168 & 107.4 & 224 & 384 & 85.8 \\
    		~ & LV-ViT-L \cite{lvvit} & 150 & 59.0 & 288 & 288 & 85.3 \\
    		~ & LV-ViT-L $\uparrow$448 \cite{lvvit} & 150 & 157.2 & 288 & 448 & 85.9 \\
    		~ & VOLO-D3 \cite{volo} & 86 & 20.6 & 224 & 224 & 85.4 \\
    		~ & VOLO-D3 $\uparrow$448 \cite{volo} & 86 & 67.9 & 224 & 448 & 86.3 \\
    		~ & \cellcolor{gray!20}{UniFormer-L$^{\star}$} &  \cellcolor{gray!20}{100} & \cellcolor{gray!20}{12.6} & \cellcolor{gray!20}{224} & \cellcolor{gray!20}{224} & \cellcolor{gray!20}{85.6} \\
            ~ & \cellcolor{gray!20}{UniFormer-L$^{\star}$ $\uparrow$384} &  \cellcolor{gray!20}{100} & \cellcolor{gray!20}{39.2} & \cellcolor{gray!20}{224} & \cellcolor{gray!20}{384} & \cellcolor{gray!20}{\textbf{86.3}} \\
    	    \Xhline{1.0pt}
    	\end{tabular}
    }
    \vspace{-0.2cm}
    \caption{\textbf{Comparison with the state-of-the-art on ImageNet.} 
    `$^{\star}$' means Token Labeling proposed in LV-ViT \cite{lvvit}. 
    `$\Upsilon$' means using RegNet-16GF \cite{regnet} as teacher.
    For UniFormer-S$\dagger$, we apply overlapped patch embedding and more blocks for fair comparison.}
    \label{results_imagenet}
\end{table}

\subsection{LightWeight UniFormer Architecture}


To build our light-weight UniFormer,
We follow most of the architectures in Section \ref{section_framework},
except that we change to use the H-UniFormer block in Stage3 and Stage4,
and adopt smaller depth, width or resolution (e.g., 128).
Specifically,
for UniFomrer-XS,
the block number,
channel number and head dimension are [3, 5, 9, 3], [64, 128, 256, 512] and 32.
For UniFomrer-XXS,
the block number,
channel number and head dimension are [2, 5, 8, 2], [56, 112, 224, 448] and 28.

Beginning from the second layer in Stage3,
we utilize the similarity score in the previous layer $\mathbf{A}^{pre}$  to guide the token shrinking.
Based on the phenomenon that the locations of the crucial tokens are basically the same among different layers~\cite{evo_vit},
we update the similarity scores via mean,
i.e., 
$\mathbf{A}=(\mathbf{A}+\mathbf{A}^{pre})/2$.
Thus it can focus on the significant tokens consistently.
By default,
we keep half of the tokens and fuse the rest (i.e., the shrinking ratio is 0.5) in our light-weight UniFormer.

\section{Experiments}
To verify the effectiveness and efficiency of our UniFormer for visual recognition,
we conduct extensive experiments on ImageNet-1K\cite{imagenet} image classification,
Kinetics-400\cite{k400}/600\cite{k600} and Something-Something V1\&V2\cite{sth} video classification,
COCO \cite{coco} object detection, instance segmentation and pose estimation,
and ADE20K \cite{ade20k} semantic segmentation.
We also perform comprehensive ablation studies to analyze each design of our UniFormer.

\begin{table}[tp]
	\centering
    \setlength\tabcolsep{0.3pt}
    \resizebox{0.99\linewidth}{!}{
    	\begin{tabular}{l|l|l|l|cc|cc}
    	    \Xhline{1.0pt}
    		\multirow{2}*{Method} & \multirow{2}*{Pretrain} & \#frame$\times$ & \makecell[c]{FLOPs} & \multicolumn{2}{c|}{K400} & \multicolumn{2}{c}{K600} \\
    		 ~ & ~ & \#crop$\times$\#clip & \makecell[c]{(G)} & Top-1 & Top-5 & Top-1 & Top-5 \\
    	    \Xhline{1.0pt}
    		SmallBig$_{EN}$\cite{smallbig}  & IN-1K & (8+32)$\times$3$\times$4 &  5700 & 78.7 & 93.7 & - & - \\
    		TDN$_{EN}$\cite{tdn}   & IN-1K & (8+16)$\times$3$\times$10 &  5940 & 79.4    & 94.4 & - & - \\
    		CT-Net$_{EN}$\cite{ct_net}  & IN-1K & (16+16)$\times$3$\times$4 &  2641 & 79.8 & 94.2 & - & -  \\
    		LGD\cite{lgd} & IN-1K & 128$\times$N/A & N/A & 79.4 & 94.4 & 81.5 & 95.6 \\
    		SlowFast\cite{slowfast} & - & 8$\times$3$\times$10 & 3180 & 77.9 & 93.2 & 80.4 & 94.8 \\
    		SlowFast+NL\cite{slowfast} & - & 16$\times$3$\times$10 & 7020 & 79.8 & 93.9 & 81.8 & 95.1 \\
    		ip-CSN\cite{csn} & Sports1M & 32$\times$3$\times$10 & 3270 & 79.2 & 93.8 & - & - \\
    		CorrNet\cite{corrnet} & Sports1M & 32$\times$3$\times$10 & 6720 & 81.0 & - & - & - \\
    		X3D-M\cite{x3d} & - & 16$\times$3$\times$10 & 186 & 76.0 & 92.3 & 78.8 & 94.5 \\
    		X3D-XL\cite{x3d} & - & 16$\times$3$\times$10 & 1452 & 79.1 & 93.9 & 81.9 & 95.5 \\
    		MoViNet-A5\cite{movienet} & - & 120$\times$1$\times$1 & 281 & 80.9 & 94.9 & 82.7 & 95.7 \\
    		MoViNet-A6\cite{movienet} & - & 120$\times$1$\times$1 & 386 & 81.5 & 95.3 & 83.5 & 96.2 \\
    	    \Xhline{1.0pt}
    		ViT-B-VTN \cite{video_transformer} & IN-21K & 250$\times$1$\times$1 & 3992 & 78.6 & 93.7 & - & - \\
    		TimeSformer-HR\cite{timesformer} & IN-21K & 16$\times$3$\times$1 & 5109 & 79.7 & 94.4 & 82.4 & 96.0 \\
    		TimeSformer-L\cite{timesformer} & IN-21K & 96$\times$3$\times$1 & 7140 & 80.7 & 94.7 & 82.2 & 95.5 \\
    		STAM \cite{stam} & IN-21K & 64$\times$1$\times$1 & 1040 & 79.2 & - & - & - \\
    		X-ViT\cite{x_vit} & IN-21K & 8$\times$3$\times$1 & 425 & 78.5 & 93.7 & 82.5 & 95.4 \\
			X-ViT\cite{x_vit} & IN-21K & 16$\times$3$\times$1 & 850 & 80.2 & 94.7 & 84.5 & 96.3 \\
			Mformer-HR\cite{motionformer} & IN-21K & 16$\times$3$\times$10 & 28764 & 81.1 & 95.2 & 82.7 & 96.1 \\
    		MViT-B,16$\times$4\cite{mvit} & - & 16$\times$1$\times$5 & 353 & 78.4 & 93.5 & 82.1 & 95.7 \\
    		MViT-B,32$\times$3\cite{mvit} & - & 32$\times$1$\times$5 & 850 & 80.2 & 94.4 & 83.4 & 96.3 \\
    		ViViT-L\cite{vivit} & IN-21K & 16$\times$3$\times$4 & 17352 & 80.6 & 94.7 & 82.5 & 95.6 \\
    		ViViT-L\cite{vivit} & JFT-300M & 16$\times$3$\times$4 & 17352 & 82.8 & 95.3 & 84.3 & 96.2 \\
    		\color{gray}{ViViT-H\cite{vivit}} & \color{gray}{JFT-300M} & \color{gray}{16$\times$3$\times$4 } & \color{gray}{99792} & \color{gray}{84.8} & \color{gray}{95.8} & \color{gray}{85.8} & \color{gray}{96.5} \\
    		Swin-T\cite{video_swin} & IN-1K & 32$\times$3$\times$4 & 1056 & 78.8 & 93.6 & - & - \\
    		Swin-B\cite{video_swin} & IN-1K & 32$\times$3$\times$4 & 3384 & 80.6 & 94.6 & - & - \\
    		Swin-B\cite{video_swin} & IN-21K & 32$\times$3$\times$4 & 3384 & 82.7 & 95.5 & 84.0 & 96.5 \\
    		\color{gray}{Swin-L-384$\uparrow$\cite{video_swin}} & \color{gray}{IN-21K} & \color{gray}{32$\times$5$\times$10 } & \color{gray}{105350} & \color{gray}{84.9} & \color{gray}{96.7} & \color{gray}{86.1} & \color{gray}{97.3} \\
    	    \Xhline{1.0pt}
    		 UniFormer-S & IN-1K & 16$\times$1$\times$4 & 167 & 80.8 & 94.7 & 82.8 & 95.8 \\
    		 UniFormer-B & IN-1K & 16$\times$1$\times$4 & 389 & 82.0 & 95.1 & 84.0 & 96.4 \\
    		 UniFormer-B & IN-1K & 32$\times$1$\times$4 & 1036 & 82.9 & 95.4 & 84.8 & 96.7 \\
    		 UniFormer-B & IN-1K & 32$\times$3$\times$4 & 3108 & \textbf{83.0} & 95.4 & \textbf{84.9} & \textbf{96.7} \\
    	    \Xhline{1.0pt}
    	\end{tabular}
    }
    \vspace{-0.2cm}
    \caption{\textbf{Comparison with the state-of-the-art on Kinetics-400\&600.}  Our UniFormer outperforms most of the current methods with much fewer computation cost.}
    \vspace{-0.5cm}
    \label{results_kinetics}
\end{table}

\subsection{Image Classification}

\textbf{Settings.} 
We train our models from scratch on the ImageNet-1K dataset \cite{imagenet}.
For a fair comparison,
we follow the same training strategy proposed in DeiT \cite{deit} by default,
including strong data augmentation and regularization.
Additionally,
we set the stochastic depth rate as 0.1/0.3/0.4 respectively for our UniFormer-S/B/L in Table \ref{model_detail}.
We train all models via AdamW \cite{adamw} optimizer with cosine learning rate schedule \cite{cosine} for 300 epochs,
while the first 5 epochs are utilized for linear warm-up \cite{warmup}.
The weight decay,
learning rate and batch size are set to 0.05, 1e-3 and 1024 respectively.
For UniFormer-S$\dagger$,
we follow state-of-the-art ViTs \cite{cswin,focal} to apply overlapped patch embedding and blocks (3/5/9/3 blocks in each stage) for fair comparisons.
As for UniFormer-B,
we use the learning rate of 8e-4 for better convergence.

\begin{table}[tp]
	\centering
    \setlength\tabcolsep{0.5pt}
    \resizebox{0.99\linewidth}{!}{
    	\begin{tabular}{l|l|l|l|cc|cc}
    	    \Xhline{1.0pt}
    		\multirow{2}*{Method} & \multirow{2}*{Pretrain} & \#frame$\times$ & \makecell[c]{FLOPs} & \multicolumn{2}{c|}{SSV1} & \multicolumn{2}{c}{SSV2} \\
    		 ~ & ~ & \#crop$\times$\#clip & \makecell[c]{(G)} & Top-1 & Top-5 & Top-1 & Top-5 \\
    	    \Xhline{1.0pt}
    		TSN\cite{tsn} & IN-1K & 16$\times$1$\times$1 & 66  & 19.9 & 47.3 & 30.0 & 60.5 \\
    		TSM\cite{tsm} &IN-1K & 16$\times$1$\times$1 & 66  & 47.2 & 77.1 & - & - \\
    		GST\cite{gst} &IN-1K & 16$\times$1$\times$1 & 59  & 48.6 & 77.9 & 62.6 & 87.9 \\
    		TEINet\cite{tei} & IN-1K & 16$\times$1$\times$1 & 66  & 49.9 & - & 62.1 & - \\
    		TEA\cite{tea} & IN-1K & 16$\times$1$\times$1 & 70  & 51.9 & 80.3 & - & - \\
    		MSNet\cite{msnet} & IN-1K & 16$\times$1$\times$1 & 101  & 52.1 & 82.3 & 64.7 & 89.4 \\
    		CT-Net\cite{ct_net} & IN-1K & 16$\times$1$\times$1 & 75  & 52.5 & 80.9 & 64.5 & 89.3 \\
    		CT-Net$_{EN}$\cite{ct_net} & IN-1K & 8+12+16+24 & 280 & 56.6 & 83.9 & 67.8 & 91.1 \\
    		TDN\cite{tdn} & IN-1K & 16$\times$1$\times$1 & 72  & 53.9 & 82.1 & 65.3 & 89.5 \\
    		TDN$_{EN}$\cite{tdn} & IN-1K & 8+16 & 198  & 56.8 & 84.1 & 68.2 & 91.6 \\
    	    \Xhline{1.0pt}
    		TimeSformer-HR\cite{timesformer} & IN-21K & 16$\times$3$\times$1 & 5109 & - & - & 62.5 & - \\
    		TimeSformer-L\cite{timesformer} & IN-21K & 96$\times$3$\times$1 & 7140 & - & - & 62.3 & - \\
    		X-ViT\cite{x_vit} & IN-21K & 16$\times$3$\times$1 & 850 & - & - & 65.2 & 90.6 \\
    		X-ViT\cite{x_vit} & IN-21K & 32$\times$3$\times$1 & 1270 & - & - & 65.4 & 90.7 \\
			Mformer-HR\cite{motionformer} & K400 & 16$\times$3$\times$1 & 2876 & - & - & 67.1 & 90.6 \\
			Mformer-L\cite{motionformer} & K400 & 32$\times$3$\times$1 & 3555 & - & - & 68.1 & 91.2 \\
    		ViViT-L\cite{vivit} & K400 & 16$\times$3$\times$4 & 11892 & - & - & 65.4 & 89.8 \\
    		MViT-B,64$\times$3\cite{mvit} & K400 & 64$\times$1$\times$3 & 1365 & - & - & 67.7 & 90.9 \\
    		MViT-B-24,32$\times$3\cite{mvit} & K600 & 32$\times$1$\times$3 & 708 & - & - & 68.7 & 91.5 \\
    		Swin-B\cite{video_swin} & K400 & 32$\times$3$\times$1 & 963 & - & - & 69.6 & 92.7 \\
    	    \Xhline{1.0pt}
    		UniFormer-S & K400 & 16$\times$1$\times$1 & 42 & 53.8 & 81.9 & 63.5 & 88.5 \\
    		UniFormer-S & K600 & 16$\times$1$\times$1 & 42 & 54.4 & 81.8 & 65.0 & 89.3 \\
    		UniFormer-S & K400 & 16$\times$3$\times$1 & 125 & 57.2 & 84.9 & 67.7 & 91.4 \\
    		UniFormer-S & K600 & 16$\times$3$\times$1 & 125 & 57.6 & 84.9 & 69.4 & 92.1 \\
    		\hline
            UniFormer-B & K400 & 16$\times$3$\times$1 & 290 & 59.1 & 86.2 & 70.4 & 92.8 \\
    		UniFormer-B & K600 & 16$\times$3$\times$1 & 290 & 58.8 & 86.5 & 70.2 & \textbf{93.0} \\
    		UniFormer-B & K400 & 32$\times$3$\times$1 & 777 & 60.9 & 87.3 & \textbf{71.2} & 92.8 \\
    		UniFormer-B & K600 & 32$\times$3$\times$1 & 777 & \textbf{61.0} & \textbf{87.6} & \textbf{71.2} & 92.8 \\
    		\color{gray}{UniFormer-B} & \color{gray}{K400} & \color{gray}{32$\times$3$\times$2} & \color{gray}{1554} & \color{gray}{61.0} & \color{gray}{87.3} & \color{gray}{71.4} & \color{gray}{92.8} \\
    		\color{gray}{UniFormer-B} & \color{gray}{K600} & \color{gray}{32$\times$3$\times$2} & \color{gray}{1554} & \color{gray}{61.2} & \color{gray}{87.6} & \color{gray}{71.3} & \color{gray}{92.8} \\
    	    \Xhline{1.0pt}
    	\end{tabular}
    }
    \vspace{-0.2cm}
    \caption{\textbf{Comparison with the state-of-the-art on Something-Something V1\&V2.} Our UniFormer achieves new state-of-the-art performances on both datasets.}
    \label{results_sth}
\vspace{-0.3cm}
\end{table}

For training high-performance ViTs,
hard distillation \cite{deit} and Token Labeling \cite{lvvit} are proposed,
both of which are complementary to our backbones.
Since Token Labeling is more efficient,
we apply it with an extra fully connected layer and auxiliary loss,
following the settings in LV-ViT \cite{lvvit}.
Different from the training settings in DeiT,
MixUp \cite{mixup} and CutMix \cite{cutmix} are not used since they conflict with MixToken \cite{lvvit}.
The base learning rate is 1.6e-3 for the batch size of 1024 by default.
Specially,
we adopt the base learning rate of 1.2e-3 and layer scale \cite{cait} for UniFormer-L to avoid NaN loss.
When fine-tuning our models on larger resolution,
i.e.,
384$\times$384,
the weight decay,
learning rate,
batch size,
warm-up epoch and total epoch are set to 1e-8, 5e-6, 512, 5 and 30.

\textbf{Results.}
In Table \ref{results_imagenet},
we compare our UniFormer with the state-of-the-art CNNs, ViTs and their combinations.
It clearly shows that our UniFormer outperforms previous models under different computation restrictions.
For example,
our UniFormer-S$\dagger$ achieves 83.4\% top-1 accuracy with only 4.2G FLOPs,
surpassing RegNetY-4G \cite{regnet}, Swin-T \cite{swin}, CSwin-T \cite{cswin} and CoAtNet \cite{coatnet} by 3.4\%, 2.1\%, 0.7\% and 1.8\% respectively.
Though EfficientNet \cite{efficientnet} comes from extensive neural architecture search,
our UniFormer outperforms it (83.9\% \vs 83.6\%) with less computation cost (8.3G \vs 9.9G).
Furthermore,
we enhance our models with Token Labeling \cite{lvvit},
which is denoted by `$^{\star}$'.
Compared with the models training with the same settings,
our UniFormer-L achieves higher accuracy but only 21\% FLOPs of LV-ViT-M\cite{lvvit} and 61\% FLOPs of VOLO-D3\cite{volo}.
Moreover,
when fine-tuned on 384$\times$384 images,
our UniFormer-L obtains 86.3\% top-1 accuracy.
It is even better than EfficientNetV2-L \cite{efficientv2} with larger input,
demonstrating the powerful learning capacity of our UniFormer.

\begin{table*}[tp]
	\centering
    \setlength\tabcolsep{4.0pt}
    \resizebox{0.99\textwidth}{!}{
    \begin{tabular}{l|c|c|c|c|c|c|c|c|c|c|c|c|c|c}
        \Xhline{1.0pt}
        \multirow{2}{*}{Method} & \#Params & FLOPs & \multicolumn{6}{c|}{Mask R-CNN 1$\times$ schedule} & \multicolumn{6}{c}{Mask R-CNN 3$\times$ + MS schedule}\\
        ~ & (M) & (G) & $AP^b$ & $AP^b_{50}$ & $AP^b_{75}$ & $AP^m$ & $AP^m_{50}$ & $AP^m_{75}$ & $AP^b$ & $AP^b_{50}$ & $AP^b_{75}$ & $AP^m$ & $AP^m_{50}$ & $AP^m_{75}$ \\
	    \Xhline{1.0pt}
        Res50 \cite{resnet} & 44 & 260  & 38.0 & 58.6 & 41.4 & 34.4 & 55.1 & 36.7
        & 41.0 & 61.7 & 44.9 & 37.1 & 58.4 & 40.1 \\
        PVT-S \cite{pvt} & 44 & 245  & 40.4 & 62.9 & 43.8 & 37.8 & 60.1 & 40.3
        & 43.0 & 65.3 & 46.9 & 39.9 & 62.5 & 42.8 \\
        TwinsP-S \cite{twins} & 44 & 245  & 42.9 & 65.8 & 47.1 & 40.0 & 62.7 & 42.9
        & 46.8 & 69.3 & 51.8 & 42.6 & 66.3 & 46.0  \\
        Twins-S \cite{twins} & 44 & 228  & 43.4 & 66.0 & 47.3 & 40.3 & 63.2 & 43.4
        & 46.8 & 69.2 & 51.2 & 42.6 & 66.3 & 45.8 \\
        Swin-T \cite{swin} & 48 & 264  & 42.2 & 64.6 & 46.2 & 39.1 & 61.6 & 42.0
        & 46.0 & 68.2 & 50.2 & 41.6 & 65.1 & 44.8 \\
        ViL-S \cite{vil} & 45 & 218  & 44.9 & 67.1 & 49.3 & 41.0 & 64.2 & 44.1
        & 47.1 & 68.7 & 51.5 & 42.7 & 65.9 & 46.2 \\
        Focal-T \cite{focal} & 49 & 291 & 44.8 & 67.7&  49.2 & 41.0 & 64.7 
        & 44.2 & 47.2 & 69.4 & 51.9 & 42.7 & 66.5 & 45.9 \\
        \rowcolor{gray!20} 
        UniFormer-S$_{h14}$ & 41 & 269 & \textbf{45.6} & \textbf{68.1} & \textbf{49.7} & \textbf{41.6} & \textbf{64.8} & \textbf{45.0} & \textbf{48.2} & \textbf{70.4} & \textbf{52.5} & \textbf{43.4} & \textbf{67.1} & \textbf{47.0} \\
        \Xhline{1.0pt}
        Res101 \cite{resnet} & 63 & 336  & 40.4 & 61.1 & 44.2 & 36.4 & 57.7 & 38.8
        & 42.8 & 63.2 & 47.1 & 38.5 & 60.1 & 41.3\\
        X101-32 \cite{resnext} & 63 & 340  & 41.9 & 62.5 & 45.9 & 37.5 & 59.4 & 40.2
        & 44.0 & 64.4 & 48.0 & 39.2 & 61.4 & 41.9 \\
        PVT-M \cite{pvt} & 64 & 302  & 42.0 & 64.4 & 45.6 & 39.0 & 61.6 & 42.1
        & 44.2 & 66.0 & 48.2 & 40.5 & 63.1 & 43.5 \\
        TwinsP-B \cite{twins} & 64 & 302  & 44.6 & 66.7 & 48.9 & 40.9 & 63.8 & 44.2
        & 47.9 & 70.1 & 52.5 & 43.2 & 67.2 & 46.3 \\
        Twins-B \cite{twins} & 76 & 340  & 45.2 & 67.6 & 49.3 & 41.5 & 64.5 & 44.8
        & 48.0 & 69.5 & 52.7 & 43.0 & 66.8 & 46.6  \\
        Swin-S \cite{swin} & 69 & 354  & 44.8 & 66.6 & 48.9 & 40.9 & 63.4 & 44.2
        & 48.5 & 70.2 & 53.5 & 43.3 & 67.3 & 46.6 \\
        Focal-S \cite{focal} & 71 & 401 & 47.4 & 69.8 & 51.9 & 42.8 & 66.6 & 46.1 & 48.8 & 70.5 & 53.6 & 43.8 & 67.7 & 47.2 \\
        CSWin-S \cite{cswin} & 54 & 342  & \textbf{47.9} & \textbf{70.1} & \textbf{52.6} & \textbf{43.2} & \textbf{67.1} & 46.2 & 
        50.0 & 71.3 & 54.7 & 44.5 & 68.4 & 47.7 \\
        \color{gray}{Swin-B \cite{swin}} & \color{gray}{107} & \color{gray}{496} & \color{gray}{46.9} & \color{gray}{-} & \color{gray}{-} & \color{gray}{42.3} & \color{gray}{-} & \color{gray}{-}
        & \color{gray}{48.5} & \color{gray}{69.8} & \color{gray}{53.2} & \color{gray}{43.4} & \color{gray}{66.8} & \color{gray}{46.9} \\
        \color{gray}{Focal-B \cite{focal}} & \color{gray}{110} & \color{gray}{533} & \color{gray}{47.8} & \color{gray}{-} & \color{gray}{-} & \color{gray}{43.2} & \color{gray}{-} & \color{gray}{-} & \color{gray}{49.0} & \color{gray}{70.1} & \color{gray}{53.6} & \color{gray}{43.7} & \color{gray}{67.6} & \color{gray}{47.0} \\
        \rowcolor{gray!20} 
        UniFormer-B$_{h14}$ & 69 & 399 & 47.4 & 69.7 & 52.1 & 43.1 & 66.0 & \textbf{46.5} & \textbf{50.3} & \textbf{72.7} & \textbf{55.3} & \textbf{44.8} & \textbf{69.0} & \textbf{48.3} \\
	    \Xhline{1.0pt}
    \end{tabular}
    }
    \vspace{-0.2cm}
    \caption{\textbf{Object detection and instance segmentation with Mask R-CNN on COCO val2017.} The FLOPs are measured at resolution 800$\times $1280. All the models are pre-trained on ImageNet-1K \cite{imagenet}.
    `$h14$' refers to hybrid UniFormer style with window size of 14 in Stage3.}
    \label{results_detection_mask}
    \vspace{-0.3cm}
\end{table*}
\begin{table}[tp]
    \centering
    \setlength\tabcolsep{4.0pt}
    \resizebox{0.99\linewidth}{!}{
        \begin{tabular}{l|l|l|c|c}
            \Xhline{1.0pt}
                Method & Pretrain & Backbone & UCF101 & HMDB51 \\
            \Xhline{1.0pt}
                C3D\cite{c3d} & Sports-1M & ResNet18 & 85.8 & 54.9 \\
                TSN\cite{tsn} & IN1K+K400 & InceptionV2 & 91.1 & - \\
                I3D\cite{i3d} & IN1K+K400 & InceptionV2 & 95.8 & 74.8 \\
                R(2+1)D\cite{r(2+1)d} & K400 & ResNet34 & 96.8 & 74.5 \\
                TSM\cite{tsm} &  IN1K+K400 &  ResNet50 & 94.5 & 70.7 \\
                STM\cite{stm} &  IN1K+K400 &  ResNet50 & 96.2 & 72.2 \\
                TEA\cite{tea} &  IN1K+K400 &  ResNet50 & 96.9 & 73.3 \\
                CT-Net\cite{ct_net} &  IN1K+K400 &  ResNet50 & 96.2 & 73.2 \\
                TDN\cite{tdn} &  IN1K+K400 &  ResNet50 & 97.4 & 76.3 \\
                VidTr\cite{vidtr} &  IN21K+K400 &  ViT-B & 96.6 & 74.4 \\
            \hline
                UniFormer & IN1K+K400 & UniFormer-S & \textbf{98.1} & \textbf{76.9} \\
            \Xhline{1.0pt}
        \end{tabular}
    }
    \vspace{-0.2cm}
    \caption{\textbf{Comparison with the state-of-the-art on UCF101 and HMDB51.} Our UniFormer has strong generalization ability.}
    \label{results_ucf_hmdb}
\vspace{-0.4cm}
\end{table}

\subsection{Video Classification}
\textbf{Settings.}
We evaluate our UniFormer on the popular Kinetics-400 \cite{k400}, Kinetics-600 \cite{k600}, UCF101 \cite{ucf} and HMDB51 \cite{hmdb},
and we verify the transfer learning performance on temporal-related datasets Something-Something (SthSth) V1\&V2 \cite{sth}.
Our codes mainly rely on PySlowFast \cite{pyslowfast}.
For training,
we adopt the same training strategy as MViT \cite{mvit} by default,
but we do not apply random horizontal flip for SthSth.
We utilize AdamW \cite{adamw} optimizer with cosine learning rate schedule \cite{cosine} to train our video backbones.. 
The first 5 or 10 epochs are used for warm-up \cite{warmup} to overcome early optimization difficulty.
For UniFormer-S,
the warmup epoch,
total epoch,
stochastic depth rate,
weight decay
are set to 10, 110, 0.1 and 0.05 respectively for Kinetics,
5, 50, 0.3 and 0.05 respectively for SthSth,
and 5, 20, 0.2 and 0.05 for UCF101 and HMDB51.
For UniFormer-B,
all the hyper-parameters are the same unless the stochastic depth rates are doubled.
Moreover,
We linearly scale the base learning rates according to the batch size,
which are 1e-4$\cdot \frac{batchsize}{32}$ for Kinetics,
2e-4$\cdot \frac{batchsize}{32}$ for SthSth,
and 1e-5$\cdot \frac{batchsize}{32}$ for UCF101 and HMDB51.

We utilize the dense sampling strategy \cite{non_local} for Kinetics, UCF101 and HMDB51,
and uniform sampling strategy \cite{tsn} for Something-Something.
To reduce the total training cost,
we inflate the 2D convolution kernels pre-trained on ImageNet for Kinetics \cite{i3d}.
To obtain a better FLOPs-accuracy balance,
Besides,
we adopt multi-clip testing for Kinetics and multi-crop testing for Something-Something.
All scores are averaged for the final prediction.

\textbf{Results on Kinetics.}
In Table \ref{results_kinetics},
we compare our UniFormer with the state-of-the-art methods on Kinetics-400 and Kinetics-600.
The first part shows the prior works using CNN.
Compared with SlowFast \cite{slowfast} equipped with non-local blocks \cite{non_local},
our UniFormer-S$_{16f}$ requires $\mathbf{42}\times$ fewer GFLOPs but obtains 1.0\% performance gain on both datasets (80.8\% \vs 79.8\% and 82.8\% \vs 81.8\%).
Even compared with MoViNet \cite{movienet},
which is a strong CNN-based models via extensive neural architecture search,
our model achieves slightly better results (82.0\% \vs 81.5\%) with fewer input frames ($16f$$\times$$4$ vs. $120f$).
The second part lists the recent methods based on vision transformers.
With only ImageNet-1K pre-training,
UniFormer-B$_{16f}$ surpasses most existing backbones with large dataset pre-training.
For example,
compared with ViViT-L \cite{vivit} pre-trained from JFT-300M \cite{jft} and Swin-B \cite{video_swin} pre-trained from ImageNet-21K,
UniFormer-B$_{32f}$ obtains comparable performance (82.9\% \vs 82.8\% and 82.7\%) with $\mathbf{16.7\times}$ and $\mathbf{3.3\times}$ fewer computation on both Kinetics-400 and Kinetics-600.
These results demonstrate the effectiveness of our UniFormer for video.

\textbf{Results on Something-Something.}
Table \ref{results_sth} presents the results on Something-Something (SthSth) V1\&V2.
Since these datasets require robust temporal relation modeling,
it is difficult for the CNN-based methods to capture long-term dependencies,
which leads to their worse results.
On the contrary,
video transformers are good at processing long sequential data and demonstrate better transfer learning capabilities \cite{transfer},
thus they achieve higher accuracy but with large computation costs.
In contrast,
our UniFormer-S$_{16f}$ combines the advantages of both convolution and self-attention,
obtaining \textbf{54.4\%}/\textbf{65.0\%} in SthSth V1/V2 with only \textbf{42} GFLOPs. 
It also demonstrates that small model UniFormer-S benefit from larger dataset pre-training (Kinetics-400, 53.8\% \vs Kinetics-600, 54.4\%),
but large model UniFormer-B do not (Kinetics-400, 59.1\% \vs Kinetics-600, 58.8\%).
We argue that the large model is easy to converge better.
Besides,
it is worth noting that our UniFormer pre-trained from Kinetis-600 outperforms all the current methods under the same settings.
In fact,
our best model achieves the new state-of-the-art results: \textbf{61.0\%} top-1 accuracy on SthSth V1 (\textbf{4.2\%} higher than TDN$_{EN}$) \cite{tdn} and \textbf{71.2\%} top-1 accuracy on SthSth V2 (\textbf{1.6\%} higher than Swin-B \cite{video_swin}).
Such results verify its high capability of spatiotemporal learning.

\textbf{Results on UCF101 and HMDB51.}
We further verify the generalization ability on UCF101 and HMDB51.
Since these datasets are relatively small,
the performances have already saturated.
As shown in Table \ref{results_ucf_hmdb},
our UniFormer significantly outperforms the previous SOTA methods,
revealing its strong generalization ability to transfer to small datasets.

\begin{table}[tp]
	\centering
    \setlength\tabcolsep{0.2pt}
    \resizebox{1.0\linewidth}{!}{
    \begin{tabular}{l|c|c|c|c|c|c|c|c}
        \Xhline{1.0pt}
        \multirow{2}{*}{Method} & \#Params & FLOPs & \multicolumn{6}{c}{ 3$\times$ + MS schedule}\\
        ~ & (M) & (G) & $AP^b$ & $AP^b_{50}$ & $AP^b_{75}$ & $AP^m$ & $AP^m_{50}$ & $AP^m_{75}$ \\
	    \Xhline{1.0pt}
	    Res50 \cite{resnet} & 82 & 739 & 46.3 & 64.3 & 50.5 & 40.1 & 61.7 & 43.4 \\
	    DeiT \cite{deit} & 80 & 889 & 48.0 & 67.2 & 51.7 & 41.4 & 64.2 & 44.3 \\
	    Swin-T \cite{swin} & 86 & 745  & 50.5 & 69.3 & 54.9 & 43.7 & 66.6 & 47.1 \\
	    Shuffle-T \cite{shuffleTR} & 86 & 746  & 50.8 & 69.6 & 55.1 & 44.1 & 66.9 & 48.0 \\
	    Focal-T \cite{focal} & 87 & 770  & 51.5 & 70.6 & 55.9 & - & - & - \\
        \rowcolor{gray!20} 
        UniFormer-S$_{h14}$ & 79 & 747 & \textbf{52.1} & \textbf{71.1} & \textbf{56.6} & \textbf{45.2} & \textbf{68.3} & \textbf{48.9} \\
        \Xhline{1.0pt}
        X101-32 \cite{resnext} & 101 & 819  & 48.1 & 66.5 & 52.4 & 41.6 & 63.9 & 45.2 \\
        Swin-S \cite{swin} & 107 & 838  & 51.8 & 70.4 & 56.3 & 44.7 & 67.9 & 48.5 \\
        Shuffle-S \cite{shuffleTR} & 107 & 844  & 51.9 & 70.9 & 56.4 & 44.9 & 67.8 & 48.6 \\
        CSWin-S \cite{cswin} & 92 & 820  & 53.7 & 72.2 & 58.4 & 46.4 & 69.6 & \textbf{50.6} \\
        \color{gray}{Swin-B \cite{swin}} & \color{gray}{145} & \color{gray}{972}  & \color{gray}{51.9} & \color{gray}{70.9} & \color{gray}{57.0} & \color{gray}{45.3} & \color{gray}{68.5} & \color{gray}{48.9} \\
        \color{gray}{Shuffle-B \cite{shuffleTR}} & \color{gray}{145} & \color{gray}{989}  & \color{gray}{52.2} & \color{gray}{71.3} & \color{gray}{57.0} & \color{gray}{45.3} & \color{gray}{68.5} & \color{gray}{48.9} \\
        \rowcolor{gray!20} 
        UniFormer-B$_{h14}$ & 107 & 878 & \textbf{53.8} & \textbf{72.8} & \textbf{58.5} & \textbf{46.4} & \textbf{69.9} & 50.4 \\
	    \Xhline{1.0pt}
    \end{tabular}
    }
    \vspace{-0.2cm}
    \caption{\textbf{Object detection and instance segmentation with Cascade Mask R-CNN on COCO val2017.} All the models are trained with 3$\times$ multi-scale schedule.
    }
    \label{results_detection_cascade}
    \vspace{-0.5cm}
\end{table}

\subsection{Object Detection and Instance Segmentation}

\textbf{Settings.}
We benchmark our models on object detection and instance segmentation with COCO2017 \cite{coco}. 
The ImageNet-1K pre-trained models are utilized as backbones and then armed with two representative frameworks:
Mask R-CNN \cite{c-maskrcnn} and Cascade Mask R-CNN \cite{maskrcnn}. 
Our codes are mainly based on mmdetection \cite{mmdetection},
and the training strategies are the same as Swin Transformer \cite{swin}.
We adopt two training schedules:
1$\times$ schedule with 12 epochs and 3$\times$ schedule with 36 epochs. 
For the 1$\times$ schedule,
the shorter side of the image is resized to 800 while keeping the longer side no more than 1333.
As for the 3$\times$ schedule,
we apply the multi-scale training strategy to randomly resize the shorter side between 480 to 800.
Besides,
we use AdamW optimizer with the initial learning rate of 1e-4 and weight decay of 0.05.
To regularize the training,
we set the stochastic depth drop rates to 0.1/0.3 and 0.2/0.4 for our small/base models with Mask R-CNN and Cascade Mask R-CNN.

\textbf{Results.}
Table \ref{results_detection_mask} reports box mAP ($AP^b$) and mask mAP ($AP^m$) of the Mask R-CNN framework.
It shows that our UniFormer variants outperform all the CNN and Transformer backbones. 
To reduce the training cost for object detection,
we utilize a hybrid UniFomer style with a window size of 14 in Stage3 (denoted by $h14$).
Specifically,
with 1$\times$ schedule,
our UniFormer brings 7.0-7.6 points of box mAP and 6.7-7.2 mask mAP against ResNet \cite{resnet} at comparable settings.
Compared with the popular Swin Transformer \cite{swin},
our UniFormer achieves 2.6-3.4 points of box mAP and 2.2-2.5 mask mAP improvement.
Moreover,
with 3$\times$ schedule and multi-scale training,
our models still consistently surpass CNN and Transformer counterparts.
For example,
our UniFormer-B outperforms the powerful CSwin-S \cite{cswin} by +0.3 box mAP and +0.3 mask mAP,
and even better than larger backbones such as Swin-B and Focal-B \cite{focal}.
Table \ref{results_detection_cascade} reports the results with the Cascade Mask R-CNN framework. 
The consistent improvement demonstrates our stronger context modeling capacity.
\begin{table}[t]
    \centering
    \setlength\tabcolsep{9pt}
    \resizebox{\linewidth}{!}{
    \begin{tabular}[t]{l|ccc}
        \Xhline{1.0pt}
        \multirow{2}{*}{Method} & \multicolumn{3}{c}{Semantic FPN 80K}\\
         &   \#Param(M) & FLOPs(G) & mIoU(\%) \\ 
        \Xhline{1.0pt}
        Res50 \cite{resnet} & 29 & 183 & 36.7 \\
        PVT-S \cite{pvt} & 28 & 161 & 39.8 \\
        TwinsP-S \cite{twins} & 28 & 162 & 44.3 \\
        Twins-S \cite{twins} & 28 & 144 & 43.2 \\
        Swin-T \cite{swin} & 32 & 182 & 41.5 \\
        \rowcolor{gray!20} 
        UniFormer-S$_{h32}$ & 25 & 199 & \textbf{46.2} \\
        \color{gray}{UniFormer-S} & \color{gray}{25} & \color{gray}{247} & \color{gray}{46.6} \\
        \Xhline{1.0pt}
        Res101 \cite{resnet} & 48 & 260 & 38.8\\
        PVT-M \cite{pvt} & 48 & 219 & 41.6 \\
        PVT-L \cite{pvt} & 65 & 283 & 42.1 \\
        TwinsP-B \cite{twins} & 48 & 220 & 44.9 \\
        TwinsP-L \cite{twins} & 65 & 283 & 46.4 \\
        Twins-B \cite{twins} & 60 & 261 & 45.3 \\
        Swin-S \cite{swin}  & 53 & 274 & 45.2 \\
        \rowcolor{gray!20} 
        \color{gray}{Twins-L \cite{twins}}  & \color{gray}{104} & \color{gray}{404} & \color{gray}{46.7} \\
        \rowcolor{gray!20} 
        \color{gray}{Swin-B \cite{swin}}  & \color{gray}{91} & \color{gray}{422} & \color{gray}{46.0} \\
        UniFormer-B$_{h32}$ & 54 & 350
        & \textbf{47.7} \\
        \color{gray}{UniFormer-B} & \color{gray}{54} & \color{gray}{471} & \color{gray}{48.0} \\
        \Xhline{1.0pt}
    \end{tabular}
    }
    \vspace{-0.2cm}
    \caption{\textbf{Semantic segmentation with semantic FPN on ADE20K.} The FLOPs are measured at resolution 512$\times$2048.
    `$h32$' means we utilize hybrid UniFormer blocks with window size of 32 in Stage3.}
    \vspace{-0.3cm}
    \label{results_ade_fpn}
\end{table}
\begin{table}[t]
	\centering
    \setlength\tabcolsep{1.8pt}
    \resizebox{\linewidth}{!}{
    \begin{tabular}[t]{l|cccc}
        \Xhline{1.0pt}
        \multirow{2}{*}{Method} & \multicolumn{4}{c}{Upernet 160K}\\
         &  \#Param.(M) & FLOPs(G) & mIoU(\%) & MS mIoU(\%) \\ 
        \Xhline{1.0pt}
        TwinsP-S \cite{twins} & 55  & 919  & 46.2 & 47.5 \\
        Twins-S \cite{twins} & 54  & 901  & 46.2 & 47.1 \\
        Swin-T \cite{swin} & 60  & 945  & 44.5 & 45.8 \\
        Focal-T \cite{focal}  & 62  & 998 & 45.8 & 47.0 \\
        Shuffle-T \cite{shuffleTR} & 60 & 949 & 46.6 & 47.8 \\
        \rowcolor{gray!20} 
        UniFormer-S$_{h32}$ & 52 & 955 & \textbf{47.0} & \textbf{48.5} \\
        \color{gray}{UniFormer-S} & \color{gray}{52} & \color{gray}{1008} & \color{gray}{47.6} & \color{gray}{48.5} \\
        \Xhline{1.0pt}
        Res101 \cite{resnet} & 86  & 1029 & - & 44.9 \\
        TwinsP-B \cite{twins} & 74 & 977  & 47.1 & 48.4 \\
        Twins-B \cite{twins} & 89  & 1020 & 47.7 & 48.9 \\
        Swin-S \cite{swin}  & 81 & 1038 & 47.6 & 49.5 \\
        Focal-T \cite{focal} & 85 & 1130 & 48.0 & 50.0 \\
        Shuffle-S \cite{shuffleTR} & 81  & 1044 & 48.4 & 49.6 \\
        \color{gray}{Swin-B \cite{swin}} & \color{gray}{121} & \color{gray}{1188} & \color{gray}{48.1} & \color{gray}{49.7} \\
        \color{gray}{Focal-B \cite{focal}} & \color{gray}{126} & \color{gray}{1354} & \color{gray}{49.0} & \color{gray}{50.5} \\
        \color{gray}{Shuffle-B \cite{shuffleTR}} & \color{gray}{121}  & \color{gray}{1196} & \color{gray}{49.0} & \color{gray}{50.5} \\
        \rowcolor{gray!20} 
        UniFormer-B$_{h32}$ & 80 & 1106 & \textbf{49.5} & \textbf{50.7} \\
        \color{gray}{UniFormer-B} & \color{gray}{80} & \color{gray}{1227} &  \color{gray}{50.0} & \color{gray}{50.8} \\
        \Xhline{1.0pt}
    \end{tabular}
    }
    \vspace{-0.2cm}
    \caption{\textbf{Semantic segmentation with Upernet on ADE20K.} The FLOPs are measured at resolution 512$\times$2048.
    `$h32$' means we utilize hybrid UniFormer blocks with window size of 32 in Stage3.}
    \vspace{-0.4cm}
    \label{results_ade_upernet}
\end{table}

\begin{table*}[t]
	\centering
    \setlength\tabcolsep{7pt}
    \resizebox{\textwidth}{!}{
    \begin{tabular}{c|l|ccc|cccccc}
        \Xhline{1.0pt}
        Arch. & Method & Input Size & \#Param(M) & FLOPs(G) & $AP$ & $AP^{50}$ & $AP^{75}$ & $AP^{M}$ & $AP^{L}$ & $AR$ \\
        \Xhline{1.0pt}
        \multirow{4}{*}{\rotatebox{90}{CNN}} &  SimpleBaseline-R101 \cite{simplebaseline} & 256$\times$192 & 53.0 & 12.4 & 71.4 & 89.3 & 79.3 & 68.1 & 78.1 & 77.1 \\
        ~ & SimpleBaseline-R152 \cite{simplebaseline} & 256$\times$192 & 68.6 & 15.7 & 72.0 & 89.3 & 79.8 & 68.7 & 78.9 & 77.8 \\
        ~ & HRNet-W$32$ \cite{hrnet} & 256$\times$192 & 28.5 & 7.1 & 74.4 & 90.5 & 81.9 & 70.8 & 81.0 & 78.9 \\
        ~ & HRNet-W$48$ \cite{hrnet} & 256$\times$192 & 63.6 & 14.6  & 75.1 & 90.6 & 82.2 & 71.5 & 81.8 & 80.4 \\
        \hline
        \multirow{6}{*}{\rotatebox{90}{CNN+Trans}} & TransPose-H-A$6$ \cite{transpose} & 256$\times$192 & 17.5 & 21.8  & 75.8 & - & - & - & - & 80.8 \\
        ~ & TokenPose-L/D$24$ \cite{tokenpose}  & 256$\times$192 & 27.5 & 11.0  & 75.8 & 90.3 & 82.5 & 72.3 & 82.7 & 80.9 \\
        ~ & HRFormer-S \cite{hrformer}& 256$\times$192 & 7.8  & 3.3 & 74.0 & 90.2 & 81.2 & 70.4 & 80.7 & 79.4 \\
        ~ & HRFormer-B \cite{hrformer}& 256$\times$192 & 43.2 & 14.1 & 75.6 & 90.8 & 82.8 & 71.7 & 82.6 & 80.8 \\
        ~ & \cellcolor{gray!20}{UniFormer-S} & \cellcolor{gray!20}{256$\times$192} & \cellcolor{gray!20}{25.2} & \cellcolor{gray!20}{4.7} & \cellcolor{gray!20}{74.0} & \cellcolor{gray!20}{90.3} & \cellcolor{gray!20}{82.2} & \cellcolor{gray!20}{66.8} & \cellcolor{gray!20}{76.7} & \cellcolor{gray!20}{79.5} \\
        ~ & \cellcolor{gray!20}{UniFormer-B} & \cellcolor{gray!20}{256$\times$192} & \cellcolor{gray!20}{53.5} & \cellcolor{gray!20}{9.2} & \cellcolor{gray!20}{75.0} & \cellcolor{gray!20}{90.6} & \cellcolor{gray!20}{83.0} & \cellcolor{gray!20}{67.8} & \cellcolor{gray!20}{77.7} & \cellcolor{gray!20}{80.4} \\
        \Xhline{1.0pt}
        \multirow{3}{*}{\rotatebox{90}{CNN}} & SimpleBaseline-R152 \cite{simplebaseline} & 384$\times$288 & 68.6 & 35.6 & 74.3 & 89.6 & 81.1 & 70.5 & 79.7 & 79.7 \\
        ~ & HRNet-W$32$ \cite{hrnet} & 384$\times$288 & 28.5 & 16.0  & 75.8 & 90.6 & 82.7 & 71.9 & 82.8 & 81.0 \\
        ~ & HRNet-W$48$ \cite{hrnet} & 384$\times$288 & 63.6 & 32.9  & 76.3 & 90.8 & 82.9 & 72.3 & 83.4 & 81.2 \\
        \hline
        \multirow{6}{*}{\rotatebox{90}{CNN+Trans}} &
        PRTR \cite{prtr} & 384$\times$288 & 57.2 & 21.6 & 73.1 & 89.4 & 79.8 & 68.8 & 80.4 & 79.8 \\
        ~ & \color{gray}{PRTR \cite{prtr}}  & \color{gray}{512$\times$384} & \color{gray}{57.2} & \color{gray}{37.8}  & \color{gray}{73.3} & \color{gray}{89.2} & \color{gray}{79.9} & \color{gray}{69.0} & \color{gray}{80.9} & \color{gray}{80.2} \\
        ~ & HRFormer-S \cite{hrformer}& 384$\times$288 & 7.8  & 7.2 & 75.6 & 90.3 & 82.2 & 71.6 & 82.5 & 80.7 \\
        ~ & HRFormer-B \cite{hrformer}& 384$\times$288 & 43.2 & 30.7 & 77.2 & 91.0 & 83.6 & \textbf{73.2} & \textbf{84.2} & 82.0 \\
        ~ & \cellcolor{gray!20}{UniFormer-S} & \cellcolor{gray!20}{384$\times$288} & \cellcolor{gray!20}{25.2} & \cellcolor{gray!20}{11.1} & \cellcolor{gray!20}{75.9} & \cellcolor{gray!20}{90.6} & \cellcolor{gray!20}{83.4} & \cellcolor{gray!20}{68.6} & \cellcolor{gray!20}{79.0} & \cellcolor{gray!20}{81.4} \\
        ~ & \cellcolor{gray!20}{UniFormer-B} & \cellcolor{gray!20}{384$\times$288} & \cellcolor{gray!20}{53.5} & \cellcolor{gray!20}{22.1} & \cellcolor{gray!20}{76.7} & \cellcolor{gray!20}{90.8} & \cellcolor{gray!20}{84.0} & \cellcolor{gray!20}{69.3} & \cellcolor{gray!20}{79.7} & \cellcolor{gray!20}{81.9} \\
        ~ & \cellcolor{gray!20}{UniFormer-S} & \cellcolor{gray!20}{448$\times$320} & \cellcolor{gray!20}{25.2} & \cellcolor{gray!20}{14.8} & \cellcolor{gray!20}{76.2} & \cellcolor{gray!20}{90.6} & \cellcolor{gray!20}{83.2} & \cellcolor{gray!20}{68.6} & \cellcolor{gray!20}{79.4} & \cellcolor{gray!20}{81.4} \\
        ~ & \cellcolor{gray!20}{UniFormer-B} & \cellcolor{gray!20}{448$\times$320} & \cellcolor{gray!20}{53.5} & \cellcolor{gray!20}{29.6} & \cellcolor{gray!20}{\textbf{77.4}} & \cellcolor{gray!20}{\textbf{91.1}} & \cellcolor{gray!20}{\textbf{84.4}} & \cellcolor{gray!20}{70.2} & \cellcolor{gray!20}{80.6} & \cellcolor{gray!20}{\textbf{82.5}} \\
        \Xhline{1.0pt}
    \end{tabular}
    }
    \vspace{-0.2cm}
    \caption{\textbf{Human Pose estimation on COCO pose estimation val set.} All the models are pre-trained on ImageNet-1K \cite{imagenet}.}
    \vspace{-0.45cm}
    \label{results_pose}
\end{table*}

\subsection{Semantic Segmentation}

\textbf{Settings.} 
Our semantic segmentation experiments are conducted on the ADE20k \cite{ade20k} dataset and our codes are based on mmseg \cite{mmseg}.
We adopt the popular Semantic FPN \cite{fpn} and Upernet \cite{upernet} as the basic framework.
For a fair comparison,
we follow the same setting of PVT \cite{pvt} to train Semantic FPN for 80k iterations with cosine learning rate schedule \cite{cosine}.
As for Upernet,
we apply the settings of Swin Transformer \cite{swin} with 160k iteration training. 
The stochastic depth drop rates are set to 0.1/0.2 and 0.25/0.4 for small/base variants with Semantic FPN and Upernet respectively.

\textbf{Results.}
Table \ref{results_ade_fpn} and Table \ref{results_ade_upernet} report the results of different frameworks. 
It shows that with the Semantic FPN framework, 
our UniFormer-S$_{h32}$/B$_{h32}$ achieve +4.7/+2.5 higher mIoU than the Swin Transformer \cite{swin} with similar model sizes.
When equipped with the UperNet framework,
they achieve +2.5/+1.9 mIoU and +2.7/+1.2 MS mIoU improvement. 
Furthermore,
when utilizing the global MHRA,
the results are consistently improved but with a larger computation cost.
More results can be found in Table \ref{ablation_adaption_segmentation}.

\begin{table}[tp]
    \centering
    \setlength\tabcolsep{2pt}
    \resizebox{0.99\linewidth}{!}{
        \begin{tabular}{l|c|cc|c|c}
        \Xhline{1.0pt}
        \multirow{2}*{Method} & \multirow{2}*{Size} & \#Param & FLOPs & \textbf{Throughput} & ImageNet \\
         ~ & ~ & (M) & (G) & \textbf{(imaegs/s) $\uparrow$} & Top-1  \\
        \Xhline{1.0pt}
        \rowcolor{gray!20} 
        UniFormer-XXS & 128 & 10.2 & 0.43 & \textbf{12886} & \textbf{76.8} \\
        MobilelViT-XXS\cite{mobilevit} & 256 & 1.3 & 0.43 & 9742  & 68.9 \\
        EfficientNet-B0\cite{efficientnet} & 224 & 5.3 & 0.42 & 9501 & 77.1 \\
        \rowcolor{gray!20} 
        UniFormer-XXS & 160 & 10.2 & 0.67 & \textbf{9382} & \textbf{79.1} \\
        PVTv2-B0\cite{pvtv2} & 224 & 3.7 & 0.57 & 8737 & 70.7 \\
        EfficientNet-B1\cite{efficientnet} & 240 & 7.8 & 0.74 & 5820  & 79.1 \\
        \rowcolor{gray!20} 
        UniFormer-XXS & 192 & 10.2 & 0.96 & \textbf{5766} & \textbf{79.9} \\
        MobileFormer\cite{mobileformer} & 224 & 14.0 & 0.51 & 4953 & 79.3 \\
        \Xhline{1.0pt}
        MobilelViT-XS\cite{mobilevit} & 256 & 2.3 & 1.1 & 4822 & 74.6 \\
        PVTv2-B1\cite{pvtv2} & 224 & 14.0 & 2.1 & 4812 & 78.7 \\
        \rowcolor{gray!20} 
        UniFormer-XS & 192 & 16.5 & 1.4 & \textbf{4492} & \textbf{81.5} \\
        \rowcolor{gray!20} 
        UniFormer-XXS & 224 & 10.2 & 1.3 & \textbf{4446} & \textbf{80.6} \\
        EfficientNet-B2\cite{efficientnet} & 260 & 9.1 & 1.1 & 4247 & 80.1 \\
        \rowcolor{gray!20} 
        UniFormer-XS & 224 & 16.5 & 2.0 & \textbf{3506} & \textbf{82.0} \\
        MobilelViT-S\cite{mobilevit} & 256 & 5.6 & 2.0 & 3360 & 78.3 \\
        EfficientNet-B3\cite{efficientnet} & 300 & 12.2 & 1.9 & 2568 & 81.6 \\
        \Xhline{1.0pt}
    \end{tabular}
    }
    \vspace{-0.2cm}
    \caption{\textbf{Comparison with the SOTA lightweight models on ImageNet.} 
    Our efficient UniFormer achieves better accuracy-throughput trade-off.
    The throughput is measured on an NVIDIA A100 GPU using the largest possible batch size for each model.
    }
    \label{results_imagenet_light}
    \vspace{-0.4cm}
\end{table}

\subsection{Pose Estimation}

\textbf{Settings.}
We evaluate the performance of UniFormer on the COCO2017\cite{coco} human pose estimation benchmark. 
For a fair comparison with previous SOTA methods, 
we employ a single Top-Down head after our backbones. 
We follow the same training and evaluation setting of mmpose \cite{mmpose} as HRFormer \cite{hrformer}. 
In addition, 
the batch size and stochastic depth drop rates are set to 1024/256 and 0.2/0.5 for small/base variants during training.

\textbf{Results.}
Table \ref{results_pose} reports results of different input resolutions on COCO validation set. 
Compared with previous SOTA CNN models, 
our UniFormer-B surpasses  HRNet-W48 \cite{hrnet} by 0.4\% AP with fewer parameters (53.5M \vs 63.6M) and FLOPs (22.1G \vs 32.9G). 
Moreover, 
our UniFormer-B can outperform the current best approach HRFormer \cite{hrformer} by 0.2\% AP with smaller FLOPs (29.6G \vs 30.7G). 
It is worth noting that HRFormer \cite{hrformer}, 
PRTR \cite{prtr}, 
TokenPose \cite{tokenpose} and TransPose \cite{transpose} are sophisticatedly designed for pose estimation task. 
On the contrary, 
our UniFormer can outperform all of them as a simple yet effective backbone.

\begin{table}[tp]
    \centering
    \setlength\tabcolsep{2pt}
    \resizebox{0.99\linewidth}{!}{
        \begin{tabular}{l|cc|cc|c|c}
        \Xhline{1.0pt}
        \multirow{2}*{Method} & \multirow{2}*{\#frame} & \multirow{2}*{Size} & \#Param & FLOPs & \textbf{Throughput} & K400 \\
         ~ & ~ & ~ & (M) & (G) & \textbf{(videos/s) $\uparrow$} & Top-1  \\
        \Xhline{1.0pt}
        \rowcolor{gray!20} 
        UniFormer-XXS & 4 & 128 & 10.4 & 1.0 & \textbf{1878} & \textbf{63.2} \\
        \rowcolor{gray!20} 
        UniFormer-XXS & 4 & 160 & 10.4 & 1.6 & \textbf{1384} & \textbf{65.8} \\
        \rowcolor{gray!20} 
        UniFormer-XXS & 8 & 128 & 10.4 & 2.0 & \textbf{1125} & \textbf{68.3} \\
        \rowcolor{gray!20} 
        UniFormer-XXS & 8 & 160 & 10.4 & 3.3 & \textbf{679} & \textbf{71.4} \\
        \rowcolor{gray!20} 
        UniFormer-XXS & 16 & 128 & 10.4 & 4.2 & \textbf{591} & \textbf{73.3} \\
        \rowcolor{gray!20} 
        UniFormer-XXS & 16 & 160 & 10.4 & 6.9 & \textbf{367} & \textbf{75.1} \\
        MoViNet-A0\cite{movienet} & 50 & 172 & 3.1 & 2.7 & 315 & 65.8 \\
        MoViNet-A1\cite{movienet} & 50 & 172 & 4.6 & 6.0 & 167 & 72.7 \\
        \rowcolor{gray!20} 
        UniFormer-XXS & 32 & 160 & 10.4 & 15.4 & \textbf{160} & \textbf{77.9} \\
        MoViNet-A2\cite{movienet} & 50 & 224 & 4.8 & 10.3 & 103 & 75.0 \\
        \rowcolor{gray!20} 
        UniFormer-XS & 32 & 192 & 16.7 & 34.2 & \textbf{75} & \textbf{78.6} \\
        X3D-XS$^{\star}$\cite{movienet} & 4 & 182 & 3.8 & 27.4 & 39 & 69.1 \\
        MoViNet-A3\cite{movienet} & 120 & 256 & 5.3 & 56.9 & 30 & 78.2 \\
        X3D-S$^{\star}$\cite{movienet} & 13 & 182 & 3.8 & 88.7 & 18 & 73.3 \\
        \Xhline{1.0pt}
    \end{tabular}
    }
    \vspace{-0.2cm}
    \caption{\textbf{Comparison with the SOTA lightweight models on Kinetics-400.} 
    Our efficient UniFormer achieves better accuracy-throughput trade-off.
    `$^{\star}$' indicates X3D is tested with 3 crops and 10 clips. 
    The throughput is measured on an NVIDIA A100 GPU.
    }
    \label{results_k400_light}
    \vspace{-0.4cm}
\end{table}

\subsection{Light-Weight UniFormer}

\textbf{Settings.}
For the light-weight UniFormer,
we follow most of the previous settings.
Differently,
we train UniFormer-XSS and UniFormer-XS for 600 epochs on ImageNet,
since the lightweight models are difficult to converge \cite{levit,mobileformer}

\textbf{Results of classification.}
Table \ref{results_imagenet_light} represents the results on ImageNet.
We roughly divide the models according to the FLOPs:
$<$1G and 1G$-$2G.
It clearly reveals that our efficient UniFormer achieves the best accuracy-throughput trade-off under similar FLOPs.
For example,
compared with the strong CNN method EfficientNet-B3\cite{efficientnet},
our UniFormer-XS$_{192}$ obtains 1.7$\times$ higher throughput with similar performance.
Compared with SOTA MobileFormer\cite{mobileformer} that combines CNN and ViT,
our UniFormer-XXS$_{192}$ obtains 0.6\% higher accuracy with 16\% higher throughput. We further fine-tune the above models with different frames on Kinetics-400.
Results in Table \ref{results_k400_light} show that our efficient backbone surpasses the SOTA lightweight video backbones by a large margin.
Compared with MoViNet-A0\cite{movienet},
our UniFormer-XXS$_{150\times 16f}$ achieves 9.3\% higher performance with 16\% higher throughput.
While compared with X3D-S\cite{x3d},
our UniFormer-XS$_{192\times 32f}$ runs 4.2$\times$ faster with 5.4\% higher accuracy.
Note that we do not apply complicated designs as in the recent lightweight methods\cite{mobileformer,mobilevit,movienet}.
Our concise extension already shows powerful performance,
which further demonstrates the great potential of UniFormer.

\begin{table}[tp]
    \centering
    \setlength\tabcolsep{1.0pt}
    \resizebox{1.0\linewidth}{!}{
    \begin{tabular}{l|c|c|c|c|c|c|c}
        \Xhline{1.0pt}
        \multirow{2}{*}{Method} & \#Params & \multicolumn{6}{c}{Mask R-CNN 1$\times$ + MS schedule}\\
        ~ & (M) & $AP^b$ & $AP^b_{50}$ & $AP^b_{75}$ & $AP^m$ & $AP^m_{50}$ & $AP^m_{75}$ \\
        \Xhline{1.0pt}
        PVTv2-B0\cite{pvtv2} & 23.5 & 38.2 & 60.5 & 40.7 & 36.2 & 57.8 & 38.6 \\
        ResNet18\cite{resnet} & 31.2 & 34.0 & 54.0 & 36.7 & 31.2 & 51.0 & 32.7 \\
        PVTv1-Tiny\cite{pvt} & 32.9 & 36.7 & 59.2 & 39.3 & 35.1 & 56.7 & 37.3 \\
        PVTv2-B1\cite{pvtv2} & 33.7 & 41.8 & 64.3 & 45.9 & 38.8 & 61.2 & 41.6 \\
        \rowcolor{gray!20} 
        UniFormer-XXS & 29.4 & 42.8 & 65.0 & 47.0 & 39.2 & 61.7 & 42.0 \\
        \rowcolor{gray!20} 
        UniFormer-XS & 35.6 & \textbf{44.6} & \textbf{67.4} & \textbf{48.8} & \textbf{40.9} & \textbf{64.2} & \textbf{44.1} \\
        \Xhline{1.0pt}
    \end{tabular}
    }
    \vspace{-0.2cm}
    \caption{\textbf{Object detection and instance segmentation of SOTA lightweight models on COCO val2017.} 
    }
    \label{results_detection_light}
    \vspace{-0.3cm}
\end{table}
\begin{table}[t]
    \centering
    \setlength\tabcolsep{8pt}
    \resizebox{\linewidth}{!}{
    \begin{tabular}[t]{l|ccc}
        \Xhline{1.0pt}
        \multirow{2}{*}{Method} & \multicolumn{3}{c}{Semantic FPN 80K}\\
         &   \#Param(M) & FLOPs(G) & mIoU(\%) \\ 
        \Xhline{1.0pt}
        PVTv2-B0\cite{pvtv2} & 7.6 & 25.0 & 37.2 \\
        ResNet18\cite{resnet} & 15.5 & 32.2 & 32.9 \\
        PVTv1-Tiny\cite{pvt} & 17.0 & 33.2 & 35.7 \\
        PVTv2-B1\cite{pvtv2} & 17.8 & 34.2 & 42.5 \\
        \rowcolor{gray!20} 
        UniFormer-XXS & 13.5 & 29.2 & 42.3 \\
        \rowcolor{gray!20} 
        UniFormer-XS & 19.7 & 32.9 & \textbf{44.4} \\
        \Xhline{1.0pt}
    \end{tabular}
    }
    \vspace{-0.2cm}
    \caption{\textbf{Semantic segmentation of SOTA lightweight models on ADE20K.} The FLOPs are measured at resolution 512$\times$512.}
    \vspace{-0.3cm}
    \label{results_ade_fpn_light}
\end{table}

\begin{table}[tp]
	\centering
    \setlength\tabcolsep{4pt}
    \resizebox{\linewidth}{!}{
    	\begin{tabular}{cccc|cc|cc}
            \Xhline{1.0pt}
    		\multirow{2}*{FFN} & \multirow{2}*{DPE} & \multicolumn{2}{c|}{MHRA} & \multicolumn{2}{c|}{ImageNet} & \multicolumn{2}{c}{K400} \\
    		~ & ~ & Size & Type & \#Param & Top-1 & GFLOPs & Top-1 \\
            \Xhline{1.0pt}
            \rowcolor{gray!20} \CheckmarkBold & \CheckmarkBold & 5 & $\mathbf{LLGG}$ & 21.5 & \textbf{82.9} & 41.8 & \textbf{79.3}  \\
    		\hline
    		\XSolidBrush & \color{lightgray}{\CheckmarkBold} & \color{gray}{5} & \color{gray}{$\mathrm{LLGG}$} & 21.3 & 82.6 & 41.0 & 78.6 \\
    		\color{lightgray}{\CheckmarkBold} & \XSolidBrush & \color{gray}{5} & \color{gray}{$\mathrm{LLGG}$} & 21.5 & 82.4 & 41.4 & 77.6 \\
    		\hline
    		\color{lightgray}{\CheckmarkBold} & \color{lightgray}{\CheckmarkBold} & 3 & \color{lightgray}{$\mathbf{LLGG}$} & 21.5 & 82.8 & 41.0 & 79.0  \\
    		\color{lightgray}{\CheckmarkBold} & \color{lightgray}{\CheckmarkBold} & 7 & \color{lightgray}{$\mathbf{LLGG}$} & 21.6 & 82.9 & 43.5 & 79.1  \\
    		\color{lightgray}{\CheckmarkBold} & \color{lightgray}{\CheckmarkBold} & 9 & \color{lightgray}{$\mathbf{LLGG}$} & 21.6 & 82.8 & 46.6 & 78.9  \\
    		\hline
    		\color{lightgray}{\CheckmarkBold} & \color{lightgray}{\CheckmarkBold} & \color{gray}{5} & $\mathbf{LLLL}$ & 23.3 & 81.9 & 31.6 & 77.2 \\
    		\color{lightgray}{\CheckmarkBold} & \color{lightgray}{\CheckmarkBold} & \color{gray}{5} & $\mathbf{LLLG}$ & 22.2 & 82.5 & 31.6 & 78.4 \\
    		\color{lightgray}{\CheckmarkBold} & \color{lightgray}{\CheckmarkBold} & \color{gray}{5} & $\mathbf{LGGG}$ & 21.6 & 82.7 & 39.0 & 79.0 \\
    		\color{lightgray}{\CheckmarkBold} & \color{lightgray}{\CheckmarkBold} & \color{gray}{5} & $\mathbf{GGGG}$ & 20.1 & 82.1 & 72.0 & 75.3 \\
            \Xhline{1.0pt}
    	\end{tabular}
    }
    \vspace{-0.2cm}
    \caption{\textbf{Structure designs.} All image models are trained for 300 epochs on ImageNet. All video models are trained for 50 epochs on Kinetics-400 with 16 frames. For fair comparisons, we guarantee the parameters and computation of all the models are similar. When modifying the stage types, we modify the stage numbers and reduce the computation of self-attention as MViT \cite{mvit} for $\mathrm{LGGG}$ and $\mathrm{GGGG}$.}
    \vspace{-0.23cm}
    \label{ablation_structure}
\end{table}

\begin{table}[tp]
    \begin{minipage}[t]{0.49\textwidth}
    \centering
    \setlength\tabcolsep{2.5pt}
    \resizebox{\linewidth}{!}{
        \begin{tabular}[t]{c|c|c|cc|c|c}
            \Xhline{1.0pt}
        	\multirow{2}*{Type} & \multirow{2}*{Joint} & \multirow{2}*{GFLOPs} & \multicolumn{2}{c|}{Pretrain} & \multicolumn{2}{c}{SSV1} \\
        	~ & ~ & ~ & Dataset & Top-1 & Top-1 & Top-5 \\
            \Xhline{0.8pt}
        	\multirow{2}*{$\mathbf{LLLL}$} &  \multirow{2}*{\CheckmarkBold} & \multirow{2}*{26.1} & IN-1K & 81.0 & 49.2 & 77.4 \\
        	~ & ~ & ~ & K400 & 77.4 & 49.2\color[RGB]{198, 40, 40}{$(+0.0)$} & 77.6\color[RGB]{17, 122, 101}{$\mathbf{(+0.2)}$} \\
        	\hline
        	\multirow{2}*{$\mathbf{LLGG}$} &  \multirow{2}*{\XSolidBrush} & \multirow{2}*{36.8} & IN-1K & 82.9 & 51.9 & 80.1 \\
        	~ & ~ & ~ & K400 & 80.1 & 51.8\color[RGB]{198, 40, 40}{$(-0.1)$} & 80.1\color[RGB]{198, 40, 40}{$(+0.0)$} \\
        	\hline
            \rowcolor{gray!20} 
            ~ & ~ & ~ & IN-1K & 82.9 & 52.0 & 80.2 \\
            \rowcolor{gray!20} 
        	\multirow{-2}*{$\mathbf{LLGG}$} &  \multirow{-2}*{\CheckmarkBold} & \multirow{-2}*{41.8} & K400 & 80.8 & \textbf{53.8}\color[RGB]{17, 122, 101}{$\mathbf{(+1.8)}$} & \textbf{81.9}\color[RGB]{17, 122, 101}{$\mathbf{(+1.7)}$} \\
            \Xhline{0.8pt}
        \end{tabular}
    }
    \vspace{-0.2cm}
    \caption{\textbf{Transfer learning.}
    When the model is gradually pre-trained from ImageNet to Kinetics-400,
    joint manner performs better.}
    \label{ablation_tranfer}
    \vspace{0.2cm}
    \end{minipage}
    \begin{minipage}[t]{0.49\textwidth}
    \centering
    \setlength\tabcolsep{4pt}
    \resizebox{\linewidth}{!}{
    \begin{tabular}[t]{c|c|c|cc|cc}
        \Xhline{1.0pt}
        \multirow{2}*{Dataset} & \multirow{2}*{Pretrain} & \#frame$\times$ & \multicolumn{2}{c|}{2D} & \multicolumn{2}{c}{\cellcolor{gray!20}{3D}} \\
        ~ & ~ & \#crop$\times$\#clip & Top-1 & Top-5 & \cellcolor{gray!20}{Top-1} & \cellcolor{gray!20}{Top-5} \\
        \Xhline{1.0pt}
        \multirow{2}*{K400} & \multirow{2}*{IN-1K} & 8$\times$1$\times$1 & 74.7 & \textbf{90.8} & \cellcolor{gray!20}{\textbf{74.9}} & \cellcolor{gray!20}{90.7} \\
        ~ & ~ & 8$\times$1$\times$4 & \textbf{78.5} & 93.2 & \cellcolor{gray!20}{78.4} & \cellcolor{gray!20}{\textbf{93.3}} \\
        \hline
        \multirow{4}*{SSV1} & \multirow{2}*{IN-1K} & 8$\times$1$\times$1 & 47.9 & 75.8 & \cellcolor{gray!20}{\textbf{48.3}} & \cellcolor{gray!20}{\textbf{76.1}}  \\
        ~ & ~ & 8$\times$3$\times$1 & \textbf{51.4} & 79.6 & \cellcolor{gray!20}{51.3} & \cellcolor{gray!20}{\textbf{79.7}} \\
        \hhline{~|-|-|-|-|-|-}
        ~ & \multirow{2}*{K400} & 8$\times$1$\times$1 & 47.9 & 75.6 & {\cellcolor{gray!20}}{\textbf{48.6}} & {\cellcolor{gray!20}}{\textbf{75.6}} \\
        ~ & ~ & 8$\times$3$\times$1 & 51.3 & 79.4 & \cellcolor{gray!20}{\textbf{51.5}} & \cellcolor{gray!20}{\textbf{79.5}} \\
        \Xhline{1.0pt}
    \end{tabular}
    }
    \vspace{-0.2cm}
    \caption{\textbf{Inflating methods of convolutional filters.} Inflating the filters to 3D performs better for different datasets.}
    \label{ablation_inflate}
    \end{minipage}
    \vspace{-0.5cm}
\end{table}  

\textbf{Results of dense prediction.}
We also verify the efficient UniFormer for COCO object detection and instance segmentation in Table \ref{results_detection_light},
and ADE20K semantic segmentation in Table \ref{results_ade_fpn_light}.
Our models obviously beat ResNet\cite{resnet} and PVTv2\cite{pvtv2} on these dense prediction tasks.
For example,
our UniFormer-XS brings 2.7 points of box mAP and 2.1 mask mAP against PVTv2-B1 on COCO,
and achieves +1.9 mIoU improvement on ADE20K.

\subsection{Ablation Studies}
\label{ablation_studies}
To inspect the effectiveness of UniFormer as the backbone,
we ablate each key structure design and evaluate the performance on image and video classification datasets.
Furthermore,
for video backbones,
we explore the vital designs of pre-training, training and testing.
Finally,
we demonstrate the efficiency of our adaption on downstream tasks,
and the effectiveness of H-UniFormer.

\subsubsection{Model designs for image and video backbones}
We conduct ablation studies of the vital components in Table \ref{ablation_structure}.

\textbf{FFN.}
As mentioned in Section \ref{mhra_section},
our UniFormer blocks in the shallow layers are instantiated as a transformer-style MobileNet block \cite{csn} with extra FFN as in ViT \cite{vit}.
Hence,
we first investigate its effectiveness by replacing our UniFormer blocks in the shallow layers with MobileNet blocks \cite{mobilenetv2}.
BN and GELU are added as the original paper,
but the expand ratios are set to 3 for similar parameters.
Note that the dynamic position embedding is kept for a fair comparison.
As expected,
our UniFormer outperforms such MobileNet block both in ImageNet (+0.3\%) and Kinetics-400 (+0.7\%).
It shows that,
FFN in our model can further mix token context at each position to boost classification accuracy.

\textbf{DPE.}
With dynamic position embedding,
our UniFormer obviously improves the top-1 accuracy by +0.5\% on ImageNet,
but +1.7\% on Kinetics-400.
It shows that via encoding the position information,
our DPE can maintain spatial and temporal order,
thus contributing to better representation learning,
especially for video.

\textbf{MHRA.}
In our local token affinity (Eq. \ref{local_a}),
we aggregate the context from a small local tube.
Hence,
we investigate the influence of this tube by changing the size from 3 to 9.
Results show that our network is robust to the tube size on both ImageNet and Kinetics-400.
We simply choose the kernel size of 5 for better accuracy.
More importantly,
we investigate the configuration of local and global UniFormer block stage by stage,
in order to verify the effectiveness of our network.
As shown in row1, 7-10 in Table \ref{ablation_structure},
when we only use local MHRA ($\mathrm{LLLL}$),
the computation cost is light.
However,
the accuracy is largely dropped (-1.0\% and -2.1\% on ImageNet and Kinetics-400) due to the lack of capacity for learning long-term dependency.
When we gradually replace local MHRA with global MHRA,
the accuracy becomes better as expected.
Unfortunately,
the accuracy is dramatically dropped with a heavy computation load when all the layers apply global MHRA ($\mathrm{GGGG}$),
i.e.,
-4.0\% on Kinetics-400.
It is mainly because that,
without local MHRA,
the network lacks the ability to extract detailed representations,
leading to severe model overfitting with redundant attention for sequential video data.

\begin{figure*}[tp]
    \vspace{-0.3cm}
    \begin{minipage}[t]{0.49\textwidth}
        \centering
        \includegraphics[width=0.99\linewidth]{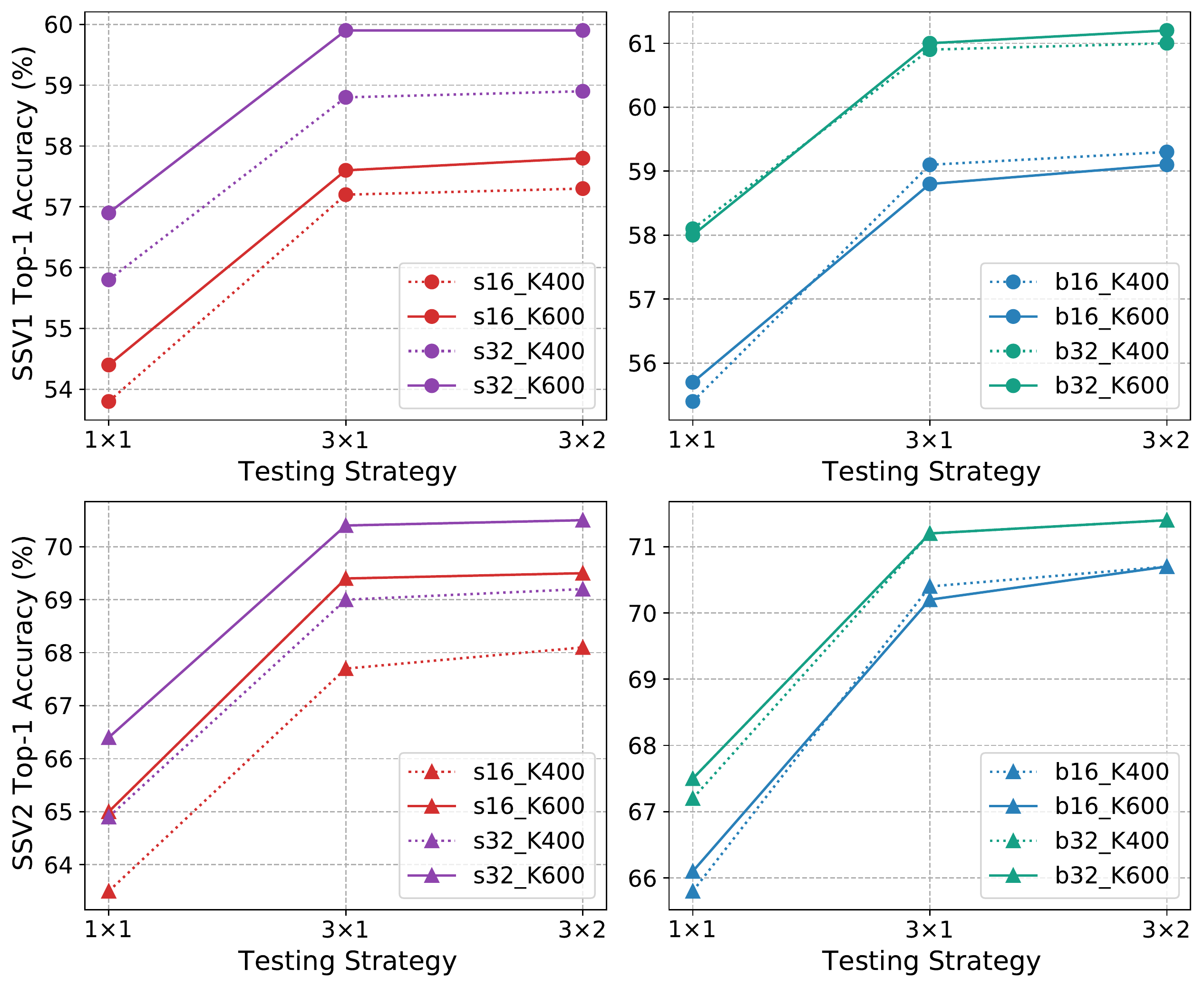}
        \vspace{-0.2cm}
        \caption{\textbf{Pre-trained dataset scales.} Small models are eager for larger dataset pre-training on both Something-Something V1 and V2.}
        \label{fig:sth_result}
    \end{minipage}
    \hspace{2mm}
    \begin{minipage}[t]{0.49\textwidth}
        \centering
        \includegraphics[width=0.99\linewidth]{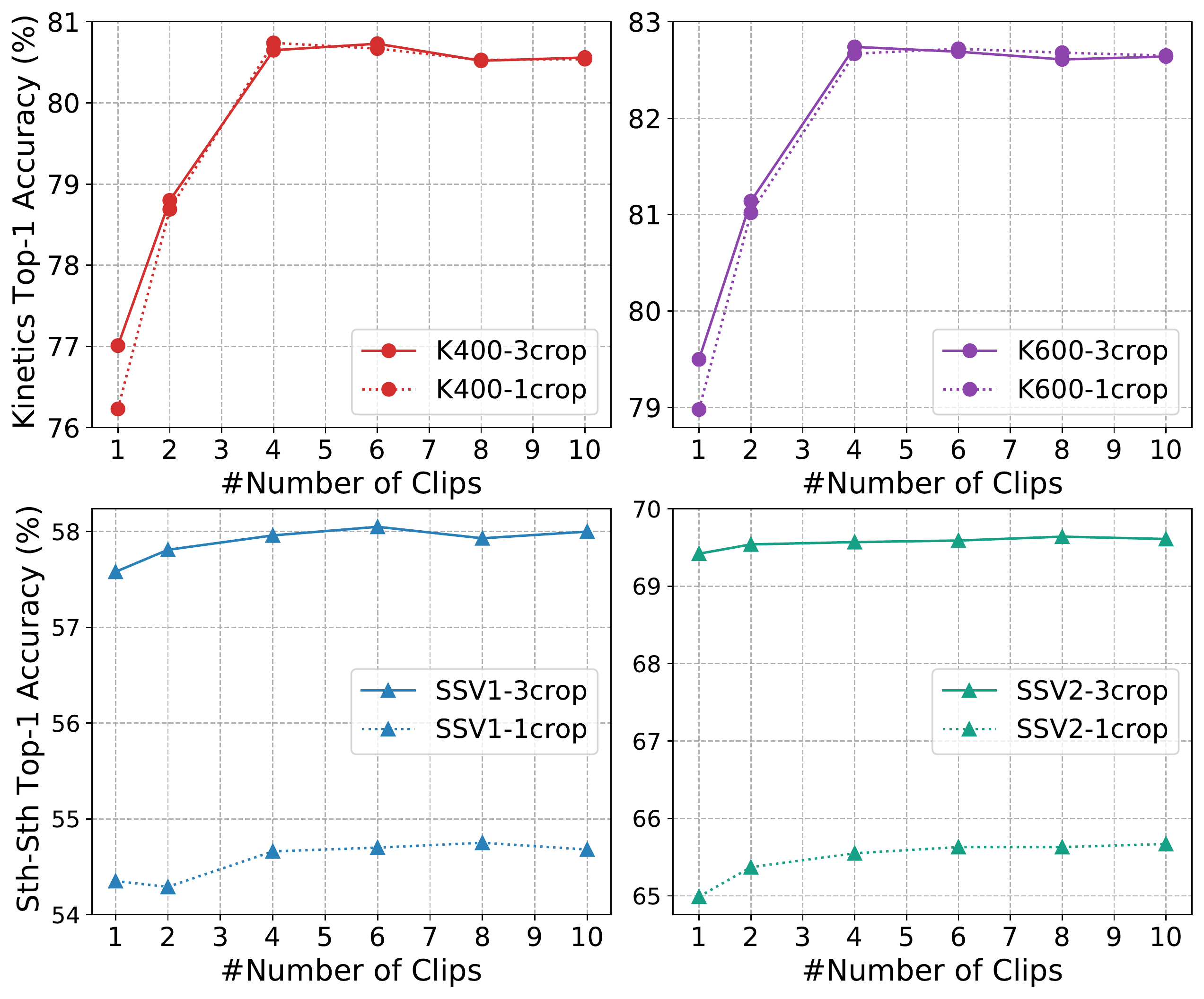}
        \vspace{-0.2cm}
        \caption{\textbf{Multi-clip/crop testing strategies.} Multi-clip testing is better for Kinetics and multi-crop testing is better for Something-Something.}
        \label{fig:test}
    \end{minipage}
    \vspace{-0.4cm}
\end{figure*}
\begin{table}[tp]
	\centering
    \setlength\tabcolsep{2.0pt}
    \resizebox{\linewidth}{!}{
    	\begin{tabular}{c|c|c|c|cc|cc}
        \Xhline{1.0pt}
    		\multirow{2}*{Model} & \#frame$\times$ & FLOPs & Sampling & \multicolumn{2}{c|}{K400} & \multicolumn{2}{c}{K600} \\
    		 ~ & \#crop$\times$\#clip & (G) & Stride & Top-1 & Top-5 & Top-1 & Top-5 \\
            \Xhline{1.0pt}
    		\multirow{4}*{Small} & \multirow{2}*{16$\times$1$\times$1} & \multirow{2}*{41.8} & 4 & 76.2 & 92.2 & 79.0 & 93.6 \\
    		~ & ~ & ~ & \cellcolor{gray!20}{8} & \cellcolor{gray!20}{\textbf{78.4}} & \cellcolor{gray!20}{\textbf{92.9}} & \cellcolor{gray!20}{\textbf{80.8}} & \cellcolor{gray!20}{\textbf{94.7}} \\
            \hhline{~|-|-|-|-|-|-|-}
    		~ &  \multirow{2}*{16$\times$1$\times$4} & \multirow{2}*{167.2} & \cellcolor{gray!20}{4} & \cellcolor{gray!20}{\textbf{80.8}} & \cellcolor{gray!20}{\textbf{94.7}} & \cellcolor{gray!20}{\textbf{82.8}} & \cellcolor{gray!20}{\textbf{95.8}} \\
    		~ & ~ & ~ & 8 & 80.8 & 94.4 & 82.7 & 95.7 \\
    		\cline{2-8}
            \hline
            \multirow{4}*{Base} & \multirow{2}*{16$\times$1$\times$1} & \multirow{2}*{96.7} & 4 & 78.1 & 92.8 & 80.3 & 94.5 \\
            ~ & ~ & ~ & \cellcolor{gray!20}{8} & \cellcolor{gray!20}{\textbf{79.3}} & \cellcolor{gray!20}{\textbf{93.4}} & \cellcolor{gray!20}{\textbf{81.7}} & \cellcolor{gray!20}{\textbf{95.0}} \\
            \hhline{~|-|-|-|-|-|-|-}
    		~ & \multirow{2}*{16$\times$1$\times$4} & \multirow{2}*{386.8} & \cellcolor{gray!20}{4} & \cellcolor{gray!20}{\textbf{82.0}} & \cellcolor{gray!20}{\textbf{95.1}} & \cellcolor{gray!20}{\textbf{84.0}} & \cellcolor{gray!20}{\textbf{96.4}} \\
    		~ & ~ & ~ & 8 & 81.7 & 94.8 & 83.4 & 96.0 \\
            \hline
    		\multirow{4}*{Small} & \multirow{2}*{32$\times$1$\times$1} & \multirow{2}*{109.6} & 2 & 77.3 & 92.4 & - & - \\
    		~ & ~ & ~  & \cellcolor{gray!20}{4} & \cellcolor{gray!20}{\textbf{79.8}} & \cellcolor{gray!20}{\textbf{93.4}} & \cellcolor{gray!20}{-}  & \cellcolor{gray!20}{-} \\
    		\cline{2-8}
    		~ & \multirow{2}*{32$\times$1$\times$4} & \multirow{2}*{438.4} & 2 & 81.2 & 94.7 & - & - \\
    		~ & ~ & ~ & \cellcolor{gray!20}{4} & \cellcolor{gray!20}{\textbf{82.0}} & \cellcolor{gray!20}{\textbf{95.1}} & \cellcolor{gray!20}{-} & \cellcolor{gray!20}{-} \\
        \Xhline{1.0pt}
    	\end{tabular}
    }
    \vspace{-0.2cm}
    \caption{\textbf{Sampling stride of dense sampling.} Larger sampling stride often achieves a higher single-clip result.}
    \label{more_results_kinetics}
    \vspace{-0.4cm}
\end{table}

\subsubsection{Pre-training, training and testing for video backbone}
In this section,
we explore more designs for our video backbones.
Firstly,
to load 2D pre-trained backbones,
it is essential to determine how to inherit self-attention and inflate convolution filters.
Hence, we compare the transfer learning performance of different MHRA configurations and inflating methods of filters.
Besides,
since we use dense sampling \cite{non_local} for Kinetics,
we should confirm the appropriate sampling stride.
Furthermore,
as we utilize Kinetics pre-trained models for SthSth,
it is interesting to explore the effect of sampling methods and dataset scales for pre-trained models.
Finally,
we ablate the testing strategies for different datasets.

\textbf{Transfer learning.}
Table \ref{ablation_tranfer} presents the results of transfer learning.
All models share the same stage numbers but the stage types are different.
For Kinetics-400,
it clearly shows that the joint version is more powerful than the separate one,
verifying that joint spatiotemporal attention can learn more discriminative video representations.
As for SthSth V1,
when the model is gradually pre-trained from ImageNet to Kinetics-400,
the performance of our joint version becomes better.
Compared with pre-training from ImgeNet,
pre-training from Kinetics-400 will further improve the top-1 accuracy by +1.8\%.
However,
such distinct characteristic is not observed in the pure local MHRA structure ($\mathrm{LLLL}$) and UniFormer with divided spatiotemporal attention.
It demonstrates that the joint learning manner is preferable for transfer learning,
thus we adopt it by default.

\begin{table}[tp]
	\centering
    \centering
    \setlength\tabcolsep{4.0pt}
    \resizebox{\linewidth}{!}{
    \begin{tabular}[t]{c|c|c|cc|cc}
        \Xhline{0.8pt}
        \multirow{2}*{Model} & Train & Pre-train & \multicolumn{2}{c|}{1crop$\times$1clip} & \multicolumn{2}{c}{3crops$\times$1clip} \\
        ~ & \#frame & \#frame$\times$\#stride & Top-1 & Top-5 & Top-1 & Top-5 \\
        \Xhline{0.8pt}
        \multirow{2}*{Small} & \multirow{2}*{16} & \cellcolor{gray!20}{16$\times$4} & \cellcolor{gray!20}{\textbf{53.8}} & \cellcolor{gray!20}{\textbf{81.9}} & \cellcolor{gray!20}{57.2} & \cellcolor{gray!20}{\textbf{84.9}} \\
        ~ & ~ & 16$\times$8 & 53.7 & 81.3 & \textbf{57.3} & 84.6 \\
        \hline
        \multirow{2}*{Base} & \multirow{2}*{16} & \cellcolor{gray!20}{16$\times$4} & \cellcolor{gray!20}{55.4} & \cellcolor{gray!20}{82.9} & \cellcolor{gray!20}{\textbf{59.1}} & \cellcolor{gray!20}{\textbf{86.2}} \\
        ~ & ~ & 16$\times$8 & \textbf{55.5} & \textbf{83.1} & 58.8 & \textbf{86.2} \\
        \hline
        \multirow{3}*{Small} & \multirow{3}*{32} & \cellcolor{gray!20}{16$\times$4} & \cellcolor{gray!20}{55.8} & \cellcolor{gray!20}{\textbf{83.6}} & \cellcolor{gray!20}{58.8} & \cellcolor{gray!20}{\textbf{86.4}} \\
        ~ & ~ & 32$\times$2 & 55.6 & 83.1 & 58.6 & 85.6 \\
        ~ & ~ & 32$\times$4 & \textbf{55.9} & 82.9 & \textbf{58.9} & 86.0 \\
        \Xhline{0.8pt}
    \end{tabular}
    }
    \vspace{-0.2cm}
    \caption{\textbf{Sampling methods of Kinetics pre-trained model.} 
    All models are pre-trained on K400 and fine-tuned on SthSth V1.}
    \label{ablation_pretrain}
    \vspace{-0.4cm}
\end{table}  

\textbf{Inflating methods.}
As indicated in I3D \cite{i3d},
we can inflate the 2D convolutional filters for easier optimization.
Here we consider whether or not to inflate the filters.
Note that the first convolutional filter in the patch stem is always inflated for temporal downsampling.
As shown in Table \ref{ablation_inflate},
inflating the filters to 3D achieves similar results on Kinetics-400,
but obtains performance improvement on SthSth V1.
We argue that Kinetics-400 is a scene-related dataset,
thus 2D convolution is enough to recognize the action.
In contrast,
SthSth V1 is a temporal-related dataset,
which requires powerful spatiotemporal modeling. 
Hence,
we inflate all the convolutional filters to 3D for better generality by default.

\textbf{Sampling stride.}
For dense sampling strategy,
the basic hyperparameter is the sampling stride of frames. 
Intuitively,
a larger sampling stride will cover a longer frame range,
which is essential for better video understanding.
In Table \ref{more_results_kinetics},
we show more results on Kinetics under different sampling strides.
As expected,
larger sampling stride (i.e. sparser sampling) often achieves higher single-clip results.
However,
when testing with multi clips,
sampling with a frame stride of 4 always performs better.

\begin{figure*}[tp]
    \vspace{-0.2cm}
    \begin{minipage}[t]{0.527\textwidth}
        \centering
        \includegraphics[width=0.99\textwidth]{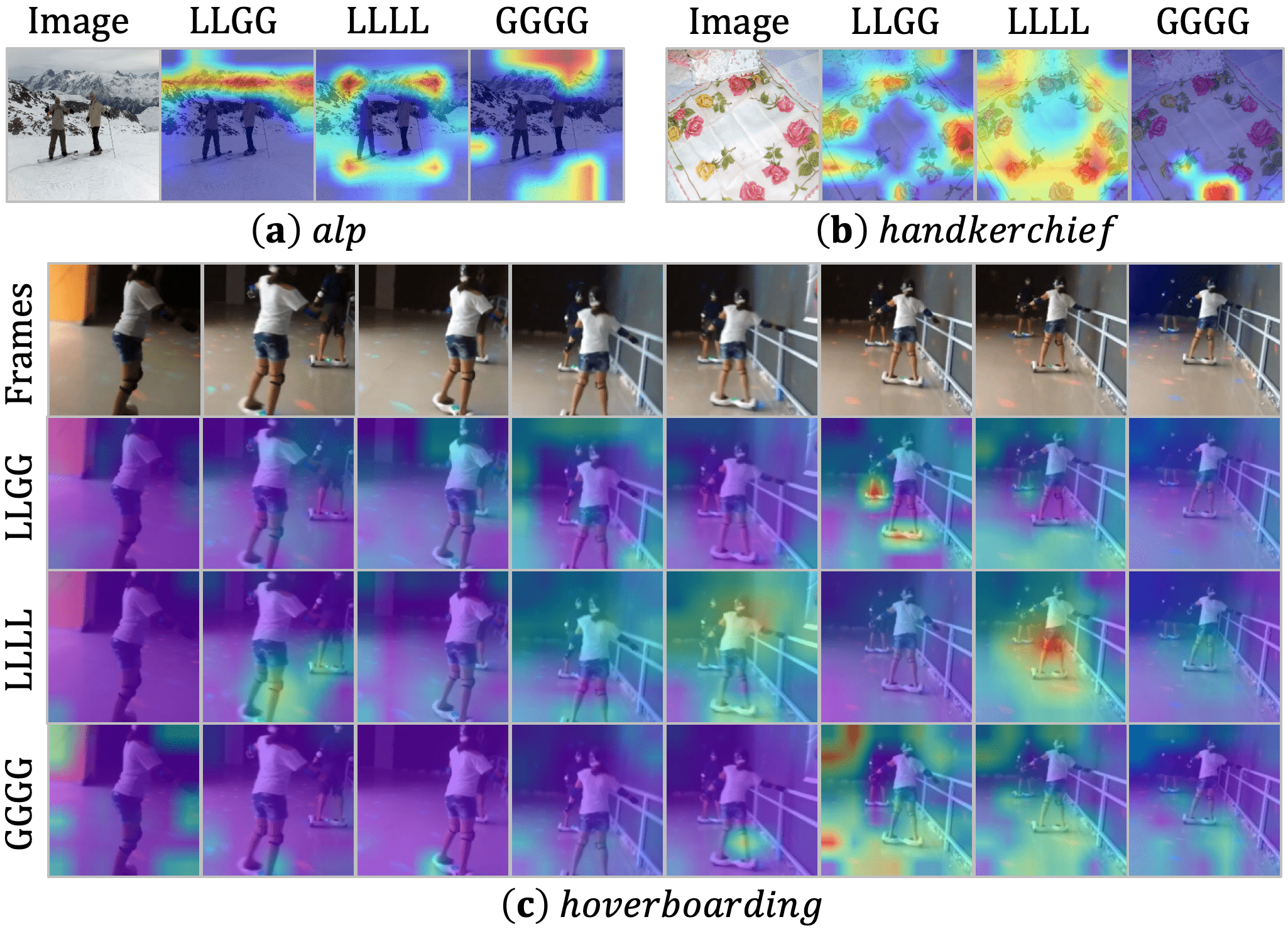}
        \vspace{-0.2cm}
        \caption{Attention visualization of different structures.}
        \label{fig:vis_image_video}
    \end{minipage}
    \begin{minipage}[t]{0.467\textwidth}
        \centering
        \includegraphics[width=0.99\textwidth]{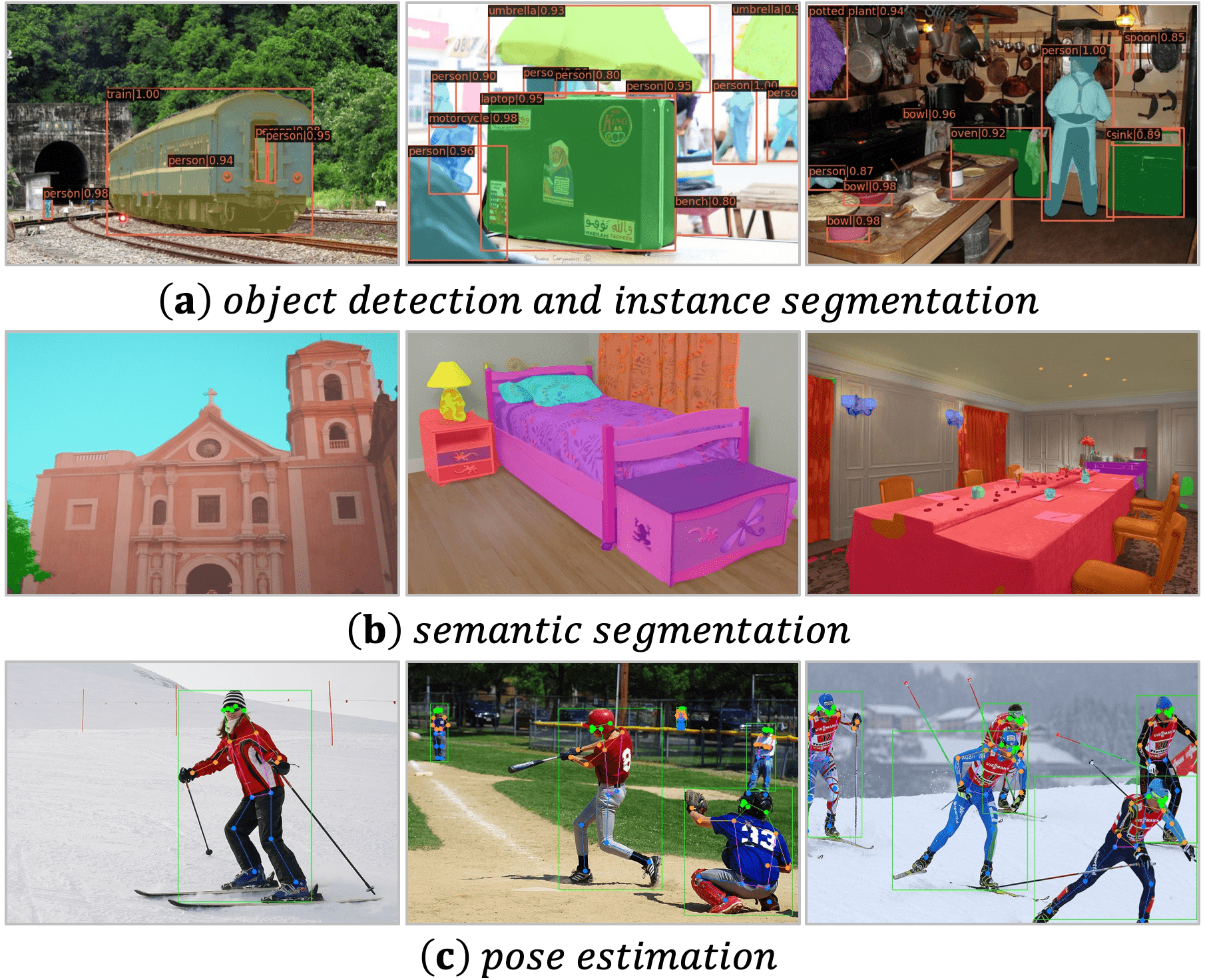}
        \vspace{-0.62cm}
        \caption{Qualitative examples for different downstream tasks.}
        \label{fig:vis_results}
    \end{minipage}
    \vspace{-0.4cm}
\end{figure*}

\begin{table}[t]
	\begin{minipage}[t]{\linewidth}
        \centering
        \setlength\tabcolsep{2.0pt}
        \resizebox{\linewidth}{!}{
            \begin{tabular}[t]{c|c|cccc|cccc}
                \Xhline{1.0pt}
                \multirow{2}*{Type} & FLOPs & \multicolumn{4}{c|}{1$\times$ + MS} & \multicolumn{4}{c}{3$\times$ + MS}\\
                ~ & (G) & $AP^b$ & $AP^b_{50}$ & $AP^m$ & $AP^m_{50}$ & $AP^b$ & $AP^b_{50}$ & $AP^m$ & $AP^m_{50}$ \\
                \Xhline{1.0pt}
                W-14 & 250 & 45.0 & 67.8 & 40.8 & 64.7 & 47.5 & 69.8 & 43.0 & 66.7 \\
                \rowcolor{gray!20} 
                H-14 & 269 & 45.4 & 68.2 & 41.4 & 64.9 & \textbf{48.2} & \textbf{70.4} & \textbf{43.4} & \textbf{67.1} \\
                G & 326 & \textbf{45.8} & \textbf{68.7} & \textbf{41.5} & \textbf{50.5} & 48.1 & 70.1 &  \textbf{43.4} & \textbf{67.1} \\
                \Xhline{1.0pt}
            \end{tabular}
        }
        \vspace{-0.2cm}
        \caption{Adaption types for object detection.}
        \label{ablation_adaption_detection}
        \vspace{0.1cm}
    \end{minipage}
	\begin{minipage}[t]{1\linewidth}
        \centering
        \setlength\tabcolsep{4.3pt}
        \resizebox{\linewidth}{!}{
            \begin{tabular}[t]{c|c|cc|cc}
                \Xhline{1.0pt}
                \multirow{2}*{Model} & \multirow{2}*{Type}  & \multicolumn{2}{c|}{Semantic FPN} & \multicolumn{2}{c}{UperNet}\\
                ~ & ~ & GFLOPs & mIoU(\%) & GFLOPs & (MS)mIoU(\%) \\
                \Xhline{1.0pt}
                \multirow{3}*{Small} & W-32 & 183 & 45.2 & 939 & (48.4)46.6 \\
                ~ & \cellcolor{gray!20}{H-32} & \cellcolor{gray!20}{199} & \cellcolor{gray!20}{46.2} & \cellcolor{gray!20}{955} & \cellcolor{gray!20}{(\textbf{48.5})47.0}  \\
                ~ & G & 247 & \textbf{46.6}& 1004 & \textbf{(48.5)47.6}  \\
                \hline
                \multirow{4}*{Base} & W-32 & 310 & 47.2 & 1066 & (50.6)49.1 \\
                ~ & \cellcolor{gray!20}{H-32} & \cellcolor{gray!20}{350} & \cellcolor{gray!20}{47.7} & \cellcolor{gray!20}{1106} & \cellcolor{gray!20}{(50.7)49.5}  \\
                ~ & G & 471 & \textbf{48.0} & 1227 & \textbf{(50.8)50.0}  \\
                \Xhline{1.0pt}
            \end{tabular}
        }
        \vspace{-0.2cm}
        \caption{Adaption types for semantic segmentation.}
        \label{ablation_adaption_segmentation}
        \vspace{0.1cm}
    \end{minipage}
	\begin{minipage}[t]{1\linewidth}
        \centering
        \setlength\tabcolsep{2.5pt}
        \resizebox{\linewidth}{!}{
            \begin{tabular}[t]{c|c|c|cccccc}
                \Xhline{1.0pt}
                \multirow{2}*{Type} & Input & FLOPs & \multirow{2}*{$AP$} & \multirow{2}*{$AP^{50}$} & \multirow{2}*{$AP^{75}$} & \multirow{2}*{$AP^M$} & \multirow{2}*{$AP^L$} & \multirow{2}*{$AR$} \\
                ~ & Size & (G) & ~ & ~ & ~ & ~ & ~ & ~ \\
                \Xhline{1.0pt}
                W-14 & 384$\times$288 & 12.3 & \textbf{76.1} & 90.8 & 83.2 & \textbf{68.9} & \textbf{79.1} & 81.1 \\
                H-14 & 384$\times$288 & 12.0 & 75.9 & 90.7 & 83.2 & 68.6 & 78.9 & 81.0 \\
                \rowcolor{gray!20} 
                G & 384$\times$288 & 11.1 & 75.9 & 90.6 & \textbf{83.4} & 68.6 & 79.0 & \textbf{81.4} \\
                \Xhline{1.0pt}
            \end{tabular}
        }
        \vspace{-0.2cm}
        \caption{Adaption types for pose estimation. Due to zero padding, the models with window UniFormer style require more computation. }
        \label{ablation_adaption_pose}
    \end{minipage}
    \setlength\tabcolsep{1.0pt}
    \vspace{-0.4cm}
\end{table}

\textbf{Sampling methods of Kinetics pre-trained model.}
For SthSth,
we uniformly sample frames as suggested in \cite{ct_net}.
Since we load Kinetics pre-trained models for fast convergence,
it is necessary to find out whether pre-trained models that cover more frames can help fine-tuning.
Table \ref{ablation_pretrain} shows that,
different pre-trained models achieve similar performances for fine-tuning.
We apply 16$\times$4 pre-training for better generalization.

\textbf{Pre-trained dataset scales.}
In Figure \ref{fig:sth_result},
we show more results on SthSth with Kinetics-400/600 pre-training.
For UniFormer-S,
Kinetics-600 pre-training consistently performs better than Kinetics-400 pre-training,
especially for large benchmark SthSth V2.
However,
both of them achieve comparable results for UniFormer-B.
These results indicate that small models are harder to converge and eager for larger dataset pre-training,
but big models are not.

\textbf{Testing strategies.}
We evaluate our network with various numbers of clips and crops for the validation videos on different datasets.
As shown in Figure \ref{fig:test},
since Kinetics is a scene-related dataset and trained with dense sampling, 
multi-clip testing is preferable to cover more frames for boosting performance.
Alternatively,
Something-Something is a temporal-related dataset and trained with uniform sampling, 
so multi-crop testing is better for capturing the discriminative motion for boosting performance.

\subsubsection{Adaption designs for downstream tasks}
We verify the effectiveness of our adaption for dense prediction tasks in Table \ref{ablation_adaption_detection},
Table \ref{ablation_adaption_segmentation} and Table \ref{ablation_adaption_pose}.
`W', `H' and `G' refer to window,
hybrid and global UniFormer style in Stage3 respectively.
Note that the pre-trained global UniFormer block can be seen as a window UniFormer block with a large window size,
thus the minimal window size in our experiments is 224/32=14.

Table \ref{ablation_adaption_detection} shows results on object detection.
Though the hybrid style performs worse than the global style with the 1$\times$ schedule,
it achieves comparable results with the 3$\times$ schedule,
which indicates that training more epochs can narrow the performance gap.
We further conduct experiments on semantic segmentation with different model variants in Table \ref{ablation_adaption_segmentation}.
As expected,
large window size and global UniFormer blocks contribute to better performances,
especially for big models.
Moreover,
when testing with multi-scale inputs,
hybrid style with a window size of 32 obtains similar results to the global style.
As for human pose estimation (Table \ref{ablation_adaption_pose}),
due to the small input resolution, 
i.e. 384$\times$288,
utilizing window style requires more computation for zero paddings.
We simply apply global UniFormer blocks for better computation-accuracy balance.

\begin{table}[tp]
    \centering
    \setlength\tabcolsep{2.8pt}
    \resizebox{\linewidth}{!}{
        \begin{tabular}[t]{c|c|c|c|l|l}
            \Xhline{1.0pt}
            Score & Running & Shrinking & \multicolumn{3}{c}{ImageNet} \\
            Token & Mean & Ratio & GFLOPs & \multicolumn{1}{c|}{Throughput} & \multicolumn{1}{c}{Top-1} \\
            \Xhline{0.8pt}
            \rowcolor{gray!20}
            \CheckmarkBold & \CheckmarkBold & 0.5 & 0.67 & 9382 & 79.1 \\
            \hline
            \XSolidBrush & \color{lightgray}{\CheckmarkBold} & 0.5 & 0.67  & 9395\color[RGB]{17, 122, 101}{(+0.1\%)} & 78.7\color[RGB]{198, 40, 40}{(-0.4)} \\
            \hline
            \color{lightgray}{\CheckmarkBold} & \XSolidBrush & 0.5 & 0.67  & 9283\color[RGB]{17, 122, 101}{(+0.0\%)} & 78.8\color[RGB]{198, 40, 40}{(-0.3)} \\
            \hline
            \color{lightgray}{\CheckmarkBold} & \color{lightgray}{\CheckmarkBold} & 1.0  & 0.91 & 8342\color[RGB]{198, 40, 40}{(-11.1\%)} & 79.9\color[RGB]{17, 122, 101}{(+0.8)} \\
            \hline
            \color{lightgray}{\CheckmarkBold} & \color{lightgray}{\CheckmarkBold} & 0.8  & 0.82 & 8452\color[RGB]{198, 40, 40}{(-9.9\%)} & 79.8\color[RGB]{17, 122, 101}{(+0.7)} \\
            \hline
            \color{lightgray}{\CheckmarkBold} & \color{lightgray}{\CheckmarkBold} & 0.6  & 0.72 & 9094\color[RGB]{198, 40, 40}{(-3.1\%)} & 79.3\color[RGB]{17, 122, 101}{(+0.2)} \\
            \hline
            \color{lightgray}{\CheckmarkBold} & \color{lightgray}{\CheckmarkBold} & 0.4  & 0.62 & 9692\color[RGB]{17, 122, 101}{(+3.3\%)} & 78.4\color[RGB]{198, 40, 40}{(-0.7)} \\
            \hline
            \color{lightgray}{\CheckmarkBold} & \color{lightgray}{\CheckmarkBold} & 0.25  & 0.55 & 10162\color[RGB]{17, 122, 101}{(+8.3\%)} & 76.8\color[RGB]{198, 40, 40}{(-2.3)} \\
            \Xhline{0.8pt}
        \end{tabular}
    }
    \vspace{-0.2cm}
    \caption{Lightweight designs for H-UniFormer.}
    \label{ablation_light}
    \vspace{-0.3cm}
\end{table}  

\subsubsection{Designs for H-UniFormer}
We further explore the lightweight designs based on UniFormer-XXS$_{160}$ in Table \ref{ablation_light}.
Firstly,
we try to remove the score token (i.e., $\mathbf{s}$) and simply use the mean of global similarity ${\rm A}_n^{global}(\mathbf{X}, \mathbf{X})$ to measure the token importance.
Results show that the learnable score token is more helpful for token selection. 
Besides,
the running mean of the similarity score (i.e., $\mathbf{A}=(\mathbf{A}+\mathbf{A}^{pre})/2$) will improve the top-1 accuracy, 
which verifies the effectiveness of consistent important tokens.
Finally,
we ablate different shrinking ratios,
where
we use the ratio of 0.5 for a better trade-off.

\subsection{Visualizations}
In Figure \ref{fig:vis_image_video},
We apply Grad-CAM \cite{grad_cam} to show the areas of the greatest concern in the last layer.
Images are sampled from ImageNet validation set \cite{imagenet} and the video is sampled from Kintics-400 validation set \cite{k400}.
It reveals that $\mathrm{GGGG}$ struggles to focus on the key object,
i.e., 
the mountain and the skateboard, 
as it blindly compares the similarity of all tokens in all layers.
Alternatively,
$\mathrm{LLLL}$ only performs local aggregation.
Hence, 
its attention tends to be coarse and inaccurate without a global view.
Different from both cases,
our UniFormer with $\mathrm{LLGG}$ can cooperatively learn local and global contexts in a joint manner.
As a result,
it can effectively capture the most discriminative information,
by paying precise attention to the mountain and the skateboard.

In Figure \ref{fig:vis_results},
we further conduct visualization on validation datasets for various downstream tasks.
Such robust qualitative results demonstrate the effectiveness of our UniFormer backbones.

\section{Conclusion}
In this paper,
we propose a novel UniFormer for efficient visual recognition,
which can effectively unify convolution and self-attention in a concise transformer format to overcome redundancy and dependency.
We adopt local MHRA in shallow layers to largely reduce computation burden and global MHRA in deep layers to learn global token relation.
Extensive experiments demonstrate the powerful modeling capacity of our UniFormer.
Via simple yet effective adaption,
our UniFormer achieves state-of-the-art results on a broad range of vision tasks with less training cost.

{
    \small
    \bibliographystyle{plain}
    \bibliography{egbib}
}








%

\begin{IEEEbiography}[{\includegraphics[width=1in,height=1.25in,clip,keepaspectratio]{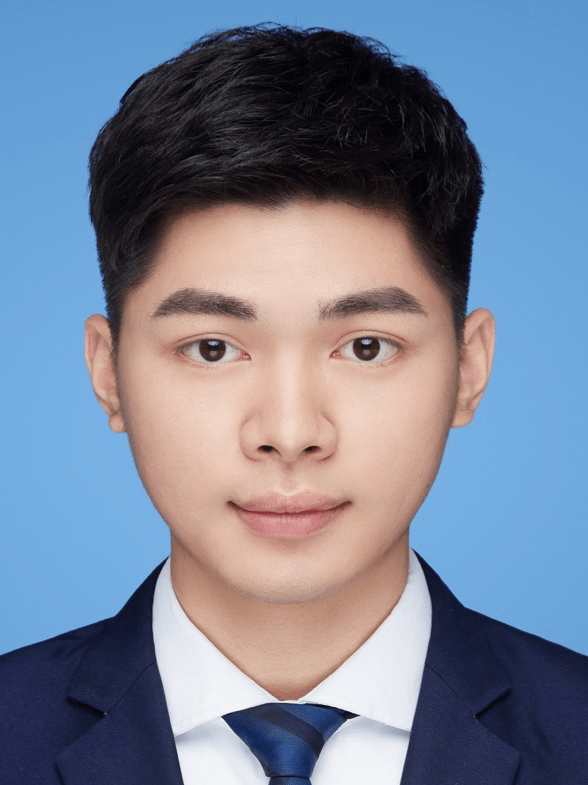}}]{Kunchang Li} is currently a second-year Ph.D. student with Shenzhen Institutes of Advanced Technology (SIAT), Chinese Academy of Science. 
He received the B.Eng. degree from Beihang University, China, in 2020. 
His research interests focus on video understanding and efficient architecture design.
\end{IEEEbiography}

\begin{IEEEbiography}[{\includegraphics[width=1in,height=1.25in,clip,keepaspectratio]{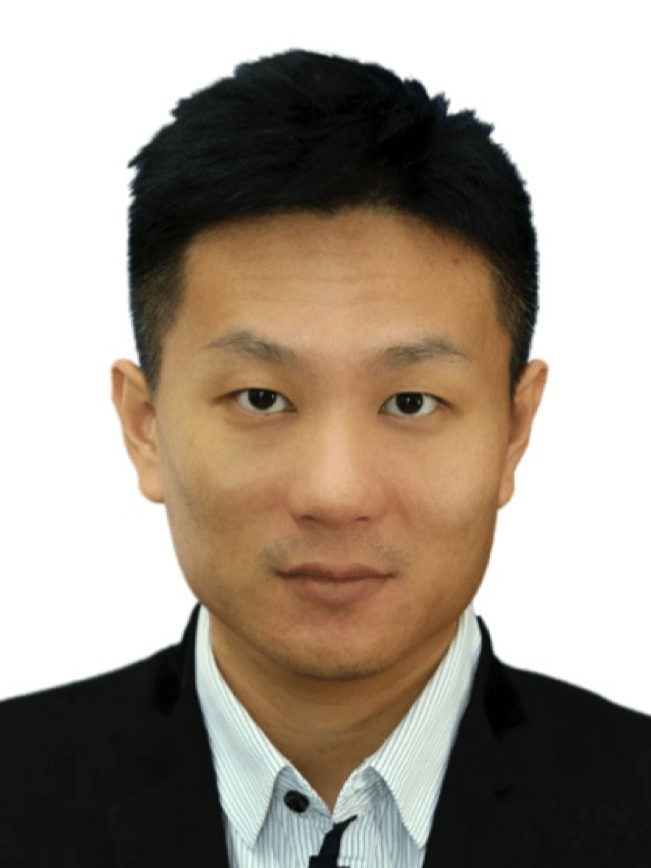}}]{Yali Wang} received the Ph.D. degree in computer science from Laval University, Quebec, QC, Canada,in 2014. 
He is currently an Associate Professor with the Shenzhen Institutes of Advanced Technology (SIAT), Chinese Academy of Sciences. 
His research interests are deep learning and computer vision, machine learning, and pattern recognition. 
\end{IEEEbiography}

\begin{IEEEbiography}[{\includegraphics[width=1in,height=1.25in,clip,keepaspectratio]{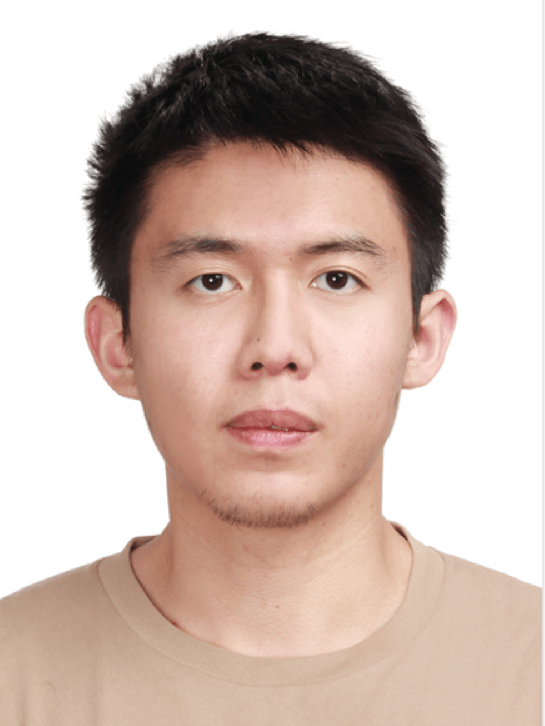}}]{Junhao Zhang} is currently a first-year Ph.D. student with the National University of Singapore, Singapore. 
He received the B.Eng. degree from Shandong University, China, in 2020. 
He was a
Research Assistant with Shenzhen Institutes of
Advanced Technology, Chinese Academy of Science.  
His research interests
are deep learning, computer vision, and robotics.
\end{IEEEbiography}

\begin{IEEEbiography}[{\includegraphics[width=1in,height=1.25in,clip,keepaspectratio]{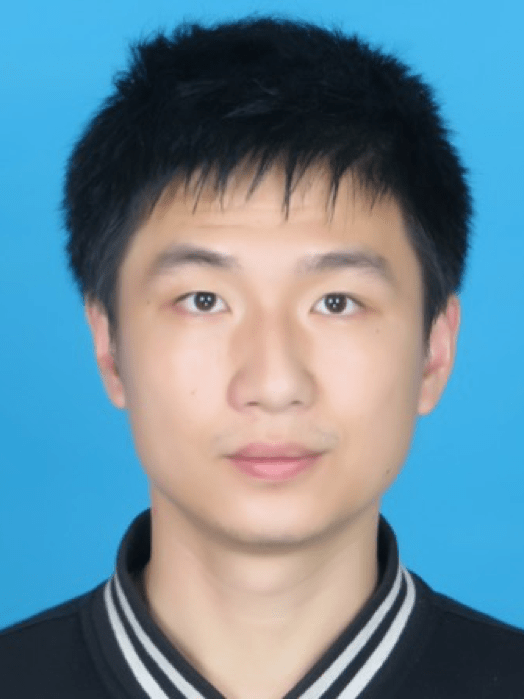}}]{Peng Gao} received his Ph.D. degree from Chinese University of Hong Kong in 2021. 
Currently, he is a Young Research Scientist at Shanghai AI Lab. 
His research interest span from efficient neural architecture design, multimodality learning and representation learning.
\end{IEEEbiography}

\begin{IEEEbiography}[{\includegraphics[width=1in,height=1.25in,clip,keepaspectratio]{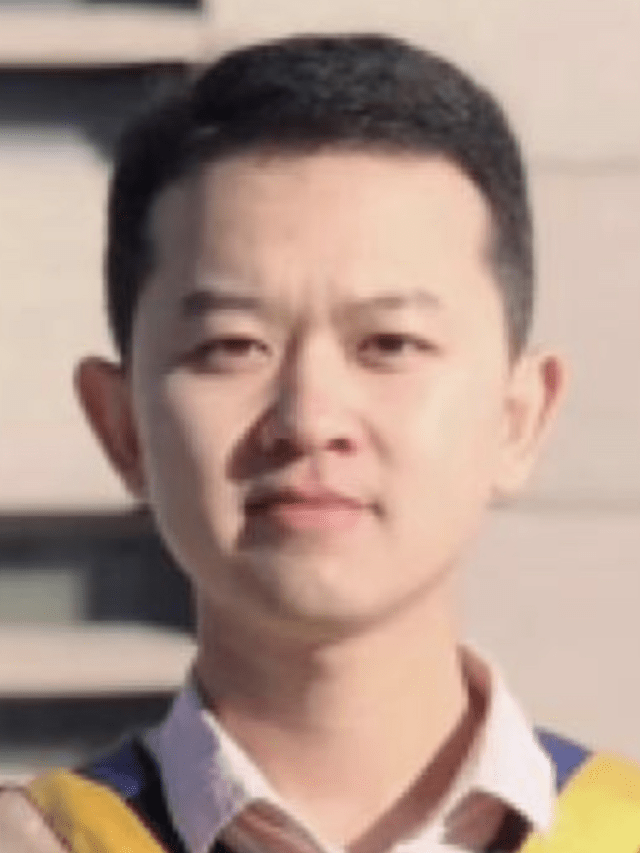}}]{Guanglu Song} is a senior researcher at SenseTime Research. 
He received a master's degree in Computer Science and Technology from Beihang University. 
His current research interests lie in computer vision, efficient architecture design, and large-scale model optimization. 
Several papers are accepted by ECCV, CVPR, ICLR, and AAAI. 
He won the championships in various famous world AI competitions such as OpenImage 2019, ActivityNet 2020, and ICCV2021-MFR. 
\end{IEEEbiography}

\begin{IEEEbiography}[{\includegraphics[width=1in,height=1.25in,clip,keepaspectratio]{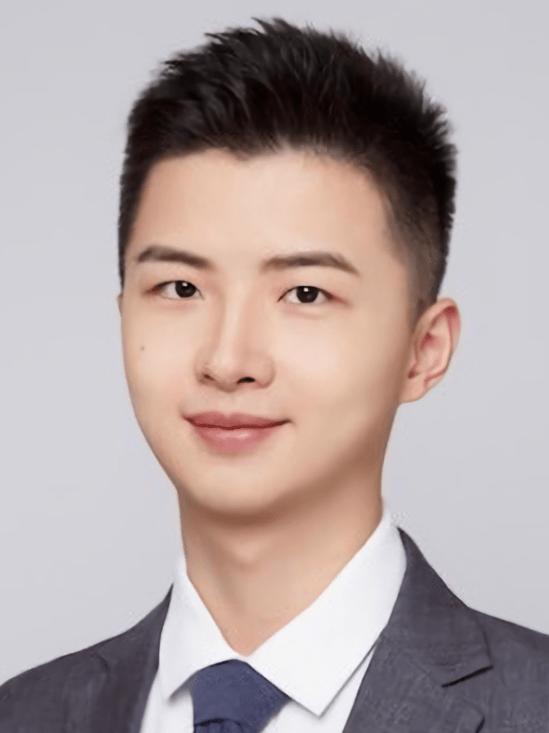}}]{Yu Liu} received his Ph.D. from the Multimedia Lab of CUHK and was the only awardee of the Google Ph.D. Fellowship in Greater China. 
He previously worked as a researcher in Microsoft Research, Google AI, and SenseTime Research. 
His research interests lie in large-scale machine learning and decision intelligence, where he published more than 30 papers with around 2000 citations. 
He won the championships in various famous world AI competitions such as ImageNet 2016, MOT 2016, OpenImage 2019, and ActivityNet 2020.
\end{IEEEbiography}

\begin{IEEEbiography}[{\includegraphics[width=1in,height=1.25in,clip,keepaspectratio]{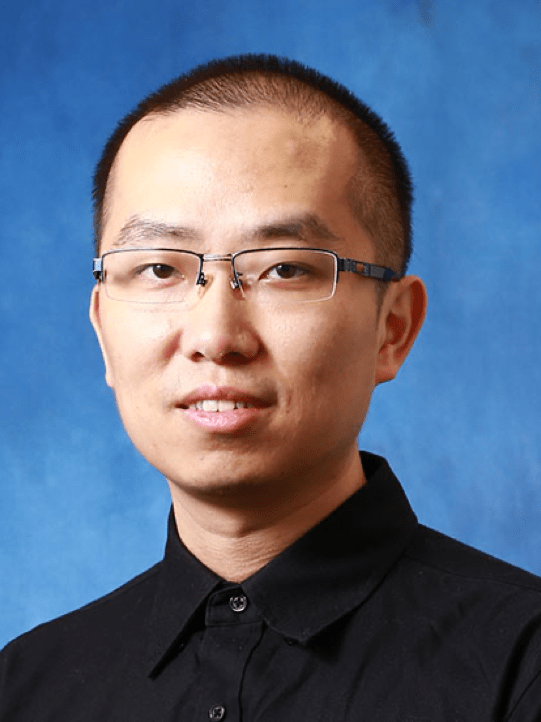}}]{Hongsheng Li} received the bachelor’s degree in automation from the East China University of Science and Technology, and the master’s and doctorate degrees in computer science from Lehigh University, Pennsylvania, in 2006, 2010, and 2012, respectively.
He is currently an assistant professor in the Department of Electronic Engineering at The Chinese University of Hong Kong. 
His research interests include computer vision, medical image analysis, and machine learning.
\end{IEEEbiography}

\begin{IEEEbiography}[{\includegraphics[width=1in,height=1.25in,clip,keepaspectratio]{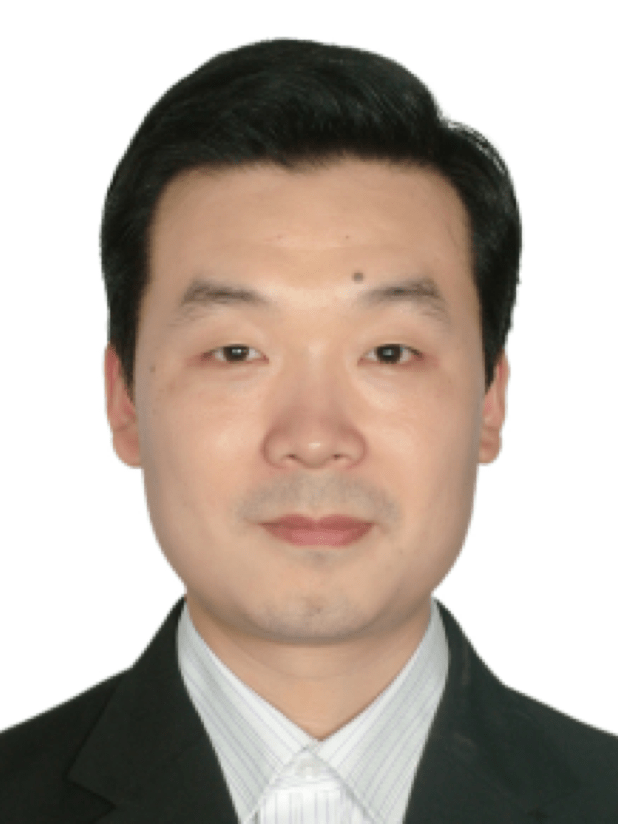}}]{Yu Qiao} (Senior Member, IEEE) is a professor with the Shenzhen Institutes of Advanced Technology (SIAT), the Chinese Academy of Science and Shanghai AI Laboratory. 
His research interests include computer vision, deep learning, and bioinformation. 
He has published more than 240 papers in international journals and conferences, including T-PAMI, IJCV, T-IP, T-SP, CVPR, ICCV etc. 
His H-index is 69, with 31,000 citations in Google scholar. 
He is a recipient of the distinguished paper award in AAAI 2021. 
His group achieved the first runner-up at the ImageNet Large Scale Visual Recognition Challenge 2015 in scene recognition, and the winner at the ActivityNet Large Scale Activity Recognition Challenge 2016 in video classification. 
He served as the program chair of IEEE ICIST 2014.
\end{IEEEbiography}




\end{document}